\documentclass[lettersize,journal]{IEEEtran}
\usepackage{amsmath,amsfonts}
\usepackage{algorithmic}
\usepackage{array}
\usepackage{textcomp}
\usepackage{stfloats}
\usepackage{url}
\usepackage{verbatim}
\usepackage{graphicx}
\def\BibTeX{{\rm B\kern-.05em{\sc i\kern-.025em b}\kern-.08em
    T\kern-.1667em\lower.7ex\hbox{E}\kern-.125emX}}
\usepackage{balance}

\usepackage{makecell}
\usepackage{float}

\usepackage{subfigure}
\usepackage[ruled,vlined,algo2e]{algorithm2e}
\usepackage{algorithm}
\usepackage{graphicx}
\usepackage{amssymb}
\allowdisplaybreaks
\usepackage{bbm}

\usepackage[numbers, sort&compress]{natbib}

\usepackage{amsthm}
\newtheoremstyle{maththeoremstyle}
  {\topsep=}
  {\topsep}
  {\it}
  {}
  {\bfseries}
  {.}
  {.5em}
  {}
\theoremstyle{maththeoremstyle}
\newtheorem{assumption}{Assumption}
\newtheorem{lemma}{Lemma}
\newtheorem{theorem}{Theorem}

\begin{document}
%
\title{Small-Scale-Fading-Aware Resource Allocation in Wireless Federated Learning}

\author{Jiacheng Wang,~\IEEEmembership{Student Member,~IEEE,}
        Le Liang,~\IEEEmembership{Member,~IEEE,}
        Hao Ye,~\IEEEmembership{Member,~IEEE,}
        \\Chongtao Guo,~\IEEEmembership{Member,~IEEE,}
        and Shi Jin,~\IEEEmembership{Fellow,~IEEE}
\thanks{Jiacheng Wang and Shi Jin are with the National Mobile Communications Research Laboratory, Southeast University, Nanjing 210096, China (e-mail: wangjiacheng@seu.edu.cn; jinshi@seu.edu.cn).}
\thanks{Le Liang is with the National Mobile Communications Research Laboratory, Southeast University, Nanjing 210096, China, and also with Purple Mountain Laboratories, Nanjing 211111, China (e-mail: lliang@seu.edu.cn).}
\thanks{Hao Ye is with the Department of Electrical and Computer Engineering, University of California, Santa Cruz, CA 95064, USA (e-mail: yehao@ucsc.edu).}
\thanks{Chongtao Guo is with the College of Electronics and Information Engineering, Shenzhen University, Shenzhen 518060, China (e-mail: ctguo@szu.edu.cn).}
}

\maketitle

\begin{abstract}

Judicious resource allocation can effectively enhance federated learning (FL) training performance in wireless networks by addressing both system and statistical heterogeneity. However, existing strategies typically rely on block fading assumptions, which overlooks rapid channel fluctuations within each round of FL gradient uploading, leading to a degradation in FL training performance. Therefore, this paper proposes a small-scale-fading-aware resource allocation strategy using a multi-agent reinforcement learning (MARL) framework. Specifically, we establish a one-step convergence bound of the FL algorithm and formulate the resource allocation problem as a decentralized partially observable Markov decision process (Dec-POMDP), which is subsequently solved using the QMIX algorithm. In our framework, each client serves as an agent that dynamically determines spectrum and power allocations within each coherence time slot, based on local observations and a reward derived from the convergence analysis. The MARL setting reduces the dimensionality of the action space and facilitates decentralized decision-making, enhancing the scalability and practicality of the solution. Experimental results demonstrate that our QMIX-based resource allocation strategy significantly outperforms baseline methods across various degrees of statistical heterogeneity. Additionally, ablation studies validate the critical importance of incorporating small-scale fading dynamics, highlighting its role in optimizing FL performance.

\end{abstract}



\begin{IEEEkeywords}
 Wireless federated learning, small-scale fading, resource allocation, multi-agent reinforcement learning.
\end{IEEEkeywords}

\IEEEpeerreviewmaketitle

\section{Introduction}

\IEEEPARstart{F}{ederated} learning (FL) is a distributed machine learning paradigm that enables collaborative model training across multiple devices without sharing raw data \cite{mcmahan2017communication}. This approach effectively mitigates the risk of data leakage and reduces storage and computational demands on central servers. In practical FL deployments, global model broadcasting and local gradient uploading occur over wireless networks \cite{bonawitz2019towards, qin2021federated}. Efficient resource management is thus essential to support FL in environments such as cellular networks, Internet of Things systems, and  heterogeneous networks \cite{yu2015multi}.

In resource-constrained wireless FL, the main challenges can be divided into two aspects: i) \textit{system heterogeneity}, which stems from the dynamic and variable network connectivity between clients and the global server \cite{ye2022decentralized}, and ii) \textit{statistical heterogeneity}, where local training data are often not independent and identically distributed (IID) \cite{Li2020On, karimireddy2020scaffold}. To be specific, the resource constraints refer to limited bandwidth and time budgets for uplink communications. However, the FL training process incurs high communication costs due to iterative model updates between clients and the central server, limiting the data that can be delivered per round for local gradient uploads. Additionally, heterogeneous and fluctuating network conditions on client devices can cause upload failures, wasting communication resources and slowing FL training \cite{yang2019scheduling}. Moreover, Non-IID local data can induce model bias, severely degrading FL convergence \cite{zhao2018federated, hsu2019measuring}. Therefore, in resource-constrained environments, an FL resource allocation strategy is necessitated to address these issues.

Existing resource allocation research in wireless FL mainly focuses on addressing system heterogeneity \cite{nishio2019client, tran2019federated, zhang2023joint}, statistical heterogeneity \cite{wang2020optimizing, cao2024fedqmix}, and approaches that concurrently manage both types of heterogeneity \cite{shi2020joint, chen2020joint, mao2024joint, zhang2022multi, zheng2024fedaeb}. Specifically, to address system heterogeneity, the FedCS protocol \cite{nishio2019client} defines and solves the client selection problem within a given time budget, which is a special case of resource allocation. To achieve more fine-grained control over resource allocation, an optimization-based strategy for communication and computing resource allocation is proposed in \cite{tran2019federated}. However, as the resource allocation problem grows more complex, solving the combinatorial optimization problem becomes increasingly challenging and difficult to implement in practice. In response, a reinforcement learning (RL)-based approach is introduced in \cite{zhang2023joint} to optimize client selection and bandwidth allocation.
While these system-heterogeneity-aware methods utilize resources effectively, they may introduce local client preferences and overlook the fairness of resource allocation, potentially undermining learning performance in non-IID settings. Regarding statistical heterogeneity, the FAVOR framework \cite{wang2020optimizing} incorporates test accuracy into reward design and employs centralized RL-based client selection. Additionally, FedQMIX \cite{cao2024fedqmix} leverages a multi-agent reinforcement learning (MARL) strategy to alleviate the dimensionality explosion faced by single-agent systems. Although these approaches have demonstrated effectiveness in handling either system or statistical heterogeneity in isolation, their performance often diminishes in environments where both types of heterogeneity coexist. To overcome this limitation, several advanced methodologies have been proposed. One strategy involves optimizing the upper bound of the FL objective via joint optimization of client selection and communication resource allocation \cite{shi2020joint, chen2020joint}. Alternatively, an extended version of the FAVOR framework integrates resource consumption into the reward function, enabling simultaneous optimization of client selection and bandwidth allocation \cite{mao2024joint}. Furthermore, incorporating a data quality factor, as introduced in \cite{zheng2024fedaeb}, effectively captures the impact of successful client uploads on FL convergence. Additionally, FedMarl \cite{zhang2022multi} adopts an MARL approach by deploying agents to individual clients, allowing autonomous determination of client participation in each FL round.

We note that, since FL typically employs neural networks with large parameter sizes, the duration of local gradient uploading often lasts far longer than the coherence time of small-scale fading in actual communication channels. In other words, channel conditions fluctuate rapidly during an FL round. However, most FL resource allocation studies rely on the block fading assumption, which fails to capture real-world channel fluctuations. This phenomenon limits the applicability of these methods \cite{liang2019spectrum, zhan2020experience}. Although the approach in \cite{chen2020joint} incorporates ergodic capacity to mitigate the impact of such fluctuations, its resource allocation decision remains static within a single FL round, lacking the adaptability in real-time fast-changing channel conditions.

In response, this paper proposes an MARL-based framework for communication resource allocation in wireless FL. This method simultaneously addresses statistical and system heterogeneity, explicitly accounting for the rapid fluctuations of small-scale fading. Specifically, it enables each client to independently determine its sub-band selection and power within each coherence time slot. This distributed decision-making effectively handles the fluctuations of small-scale fading. Additionally, the proposed MARL method effectively protects user privacy while reducing the dimensionality of the action space, thereby enhancing its applicability in practical deployments. The main contributions of this paper are summarized as follows:

\begin{itemize}
	\item We introduce a practical FL framework that explicitly accounts for rapid fluctuations in small-scale fading during the local gradient uploading process. In this framework, each client shares limited channel resources and dynamically selects its spectrum and power allocation for each time slot. At the end of each FL round, the success of the upload process is evaluated based on the sum capacity within the limited time budget.
    
    \item We establish a one-step convergence bound for the FL algorithm under mild assumptions on the local gradients, carefully considering the impact of local drift due to multiple local training in non-IID settings. This analysis elucidates how the successful transmission of local gradients contributes to the overall convergence behavior of FL, providing critical insights for developing effective resource allocation strategies.
    
    \item  We model the resource allocation problem as a decentralized partially observable Markov decision process (Dec-POMDP). In this formulation, local observations are strategically selected to ensure their availability during the decentralized execution phase. Additionally, the reward is carefully designed based on insights from the convergence analysis, effectively addressing the issue of sparse rewards from the perspective of each time slot in a single FL round.
    
    \item We solve the Dec-POMDP by developing a high-performing resource allocation strategy that accounts for small-scale fading fluctuations. Experimental results show that our proposed method significantly outperforms baseline approaches across various levels of statistic heterogeneity. Furthermore, ablation studies confirms the critical importance of integrating small-scale fading dynamics into the allocation process.
\end{itemize}

The remainder of this paper is organized as follows. Section \ref{sec:System Model} outlines the FL process and the spectrum-sharing communication model. Afterwards, Section \ref{sec:Convergence_Analysis} presents the assumptions and one-step convergence analysis of the FL framework defined in the previous section. Section \ref{sec:MARL} elaborates on our proposed resource allocation strategy, and outlines the detailed MARL formulation and the training procedure. Finally, Section \ref{sec:Simulation_Results} presents experimental results, followed by the conclusion of this paper made in Section \ref{sec:Conclusion}.

\section{System Model}\label{sec:System Model}

\subsection{Federated Learning Process}

Consider a system with $N$ FL clients, each possessing a local dataset $\mathcal{D}_n$. The local objective function for client $n$ is defined as
\begin{equation}
	F_n(w)=\frac{1}{\left|\mathcal{D}_n\right|}\sum_{x\in\mathcal{D}_n}\ell(w;x),
\end{equation}
where $\ell(w;x)$ denotes the loss function, which measures the difference between the model prediction and the ground truth for a specific training example $x$, given model parameters $w$. The objective of FL training is to minimize the global objective function $F(w)$, which represents the weighted average of the objective functions of all clients and is defined as
\begin{equation}
    F(w) = \sum_{n\in\mathcal{N}} \lambda_n F_n(w),
\end{equation}
where $\mathcal{N}=\{1,\dots,N\}$ is the set of all clients, and $\lambda_n$ is the weighting coefficient of client $n$. To simplify the derivation, we assume all weighting coefficients $\lambda_n$ are equal to ${1}/{N}$.

This paper utilizes the Federated Averaging (FedAvg) algorithm introduced by \cite{mcmahan2017communication}, where the FL server is deployed at the base station (BS) and FL clients are distributed across mobile devices within a single cell of a cellular network. Interaction between the clients and the BS is facilitated through wireless channels. A typical FL round includes the following key steps:

\begin{itemize}
\item \textbf{Global Model Broadcasting:}
The BS broadcasts the global model parameters of the $t$-th FL round, $w_t$, to all $N$ clients. Given substantial bandwidth and power of the downlink channel and the BS, this process can be assumed to be error-free \cite{tran2019federated}, and thus we have
\begin{equation}
w_{n,t}^{(0)}=w_t, \forall n\in\mathcal{N},
\end{equation}
where $w_{n,t}^{(i)}$ represents the local model parameters of the $i$-th local training epoch.

\item \textbf{Local Training:}
After receiving the model parameters, each client $n$ performs $E$ rounds of mini-batch stochastic gradient descent (SGD) in parallel, given by
\begin{equation}
    w_{n,t}^{(i)}=w_{n,t}^{(i-1)}-\eta_\text{l}\widetilde{\nabla}F_n\left(w_{n,t}^{(i-1)}\right),\forall n \in\mathcal{N},
\end{equation}
where 
\begin{equation}\label{equ:local_comp}
    \widetilde{\nabla}F_n\left(w_{n,t}^{\left(i-1\right)}\right)=\frac{1}{B_\mathcal{D}}\sum_{x\in\mathcal{B}_{n,t,i}}{\nabla \ell\left(w_{n,t}^{\left(i-1\right)};x\right)}
\end{equation}
\vspace{-0.3cm}

is the local stochastic gradient in the $i$-th local training epoch and $\eta_\text{l}$ is the local learning rate. The mini-batch $\mathcal{B}_{n,t,i}$ is uniformly and randomly sampled from the local dataset $\mathcal{D}_n$ and the batch size $B_\mathcal{D}=\left|\mathcal{B}_{n,t,i}\right|$ remains constant.

\item \textbf{Local Gradient Uploading:} 
Upon completing local training, each client $n$ uploads its cumulative stochastic gradient $\tilde{g}_{n,t}$ to the BS in the previous step, which is computed as
\vspace{-0.2cm}
\begin{equation}
\begin{aligned}
    {\tilde{g}}_{n,t}&=w_t-w_{n,t}^{(E)}\\
    &=\sum_{i=1}^{E}{\eta_\text{l}\widetilde{\nabla}F_n\left(w_{n,t}^{(i-1)}\right)}.
\end{aligned} 
\end{equation}
\vspace{-0.3cm}

However, limited bandwidth and dynamic network conditions may disrupt this process, leading to upload failures. 

\item \textbf{Global Model Update:} 
The gradients uploaded by clients are received by the server at the BS for global model aggregation. Let the binary variable $s_{n,t} \in \{0,1\}$ indicate whether the gradient from client $n$ in the $t$-th FL round is successfully uploaded. Consequently, the set of clients that successfully uploaded gradients is defined as
\begin{equation}
    \mathcal{N}_t=\{n|s_{n,t}=1,\forall n\in\mathcal{N}\}.
\end{equation}

The global model update formula is then given by
\begin{equation}
\label{equ:global_aggregation}
    w_{t+1}= w_t-\frac{\eta_\text{g}}{\left|\mathcal{N}_t\right|}\sum_{n\in \mathcal{N}_t}{\tilde{g}}_{n,t},
\end{equation}
\vspace{-0.3cm}

where $\eta_\text{g}$ is the global learning rate and $\tilde{g}_{t}$ denotes the aggregated stochastic gradient.
\end{itemize}

Through iteratively executing the above steps, the BS and FL clients collaboratively train the model and gradually minimize the global objective function. In this paper, the FL training process consists of $T$ rounds. Assume that the gradient to be uploaded contains a total of $S$ bits. Therefore, in each communication round, each client needs to upload $S$ bits within $T_s$ time slots to perform local gradient uploading.

\subsection{Spectrum-Sharing Communication Model}

We assume that all clients and the central server are equipped with a single antenna. In each time slot, clients intelligently select their sub-band and power level based on network conditions and the FL process. To fully utilize communication resources during uploading, we propose a spectrum-sharing scheme similar to those in \cite{ye2019deep, liang2019spectrum}. We assume that the band is divided into $C$ sub-bands, each with a bandwidth of $B$. The local gradient upload process consists of $T_s$ coherent time slots, with each time slot having a duration $T_d$ equal to the coherence time of small-scale fading. Therefore, the channel power gain $H_{n,t,t_s}^{(c)}$ from client $n$ to the BS in the $c$-th sub-band during the $t_s$-th time slot of the $t$-th FL round is given by
\begin{equation}
    H_{n,t,t_s}^{(c)}=\alpha_{n,t}h_{n,t,t_s}^{(c)},
\end{equation}
where $\alpha_{n,t}$ and $h_{n,t,t_s}^{(c)}$ are large-scale and small-scale fading components, respectively. Specifically, the large-scale fading $\alpha_{n,t}$ includes path loss and shadowing, which is assumed to vary in each round and remain the same across sub-bands during the gradient uploading process. In contrast, the small-scale fading $h_{n,t,t_s}^{(c)}$ continuously fluctuates in each time slot and varies across sub-bands. 

We assume that each client can occupy at most one sub-band during each time slot with duration $T_d$. Moreover, let $c_{n,t,t_s}\in\mathcal{C}=\{1,\dots,C\}$ represent the index of the sub-band occupied by the $n$-th client during the $t_s$-th time slot of the $t$-th FL round, and let $p_{n,t,t_s}$ denote the corresponding transmit power. The signal-to-interference-plus-noise ratio (SINR) from client $n$ to the BS during the $t_s$-th time slot of the $t$-th FL round is expressed as
\begin{equation}
    \text{SINR}_{n, t, t_s} =\frac{p_{n,t,t_{s}}H_{n,t,t_{s}}^{(c_{n,t,t_{s}})}}{\sigma^{2}+\sum_{n^{\prime}\in\Xi_{n,t,t_{s}}}p_{n^{\prime},t,t_{s}}H_{n^{\prime},t,t_{s}}^{(c_{n,t,t_{s}})}},
\end{equation}
where $\Xi_{n, t, t_s}$ represents the set of all other clients transmitting on the same sub-band as client $n$, i.e.,
\begin{equation}
    \Xi_{n,t,t_s}=\{n'|c_{n',t,t_s}=c_{n,t,t_s}, n'\neq n, \forall n'\in \mathcal{N}\}.
\end{equation}
The instantaneous channel capacity is then obtained as 
\begin{equation}
    C_{n,t,t_s}={B\text{log}_2(1+\text{SINR}_{n,t,t_s})}.
\end{equation}

We set a threshold for the sum capacity and ignore incomplete gradient uploads. Therefore, given the SINR for all $T_s$ time slots within the $t$-th FL round, the success of each client's gradient upload is determined as
\begin{equation}\label{equ:successful_upload_gradient}
	s_{n,t} = \mathbbm{1}\left(\sum\limits_{t_s=1}^{T_s}C_{n,t,t_s} \geq \frac{S}{ T_d}\right) \in \left\{0, 1\right\},
\end{equation}
where $\mathbbm{1}(\cdot)$ denotes the binary indicator function.



The goal of this paper is to develop a sub-band selection and power allocation strategy within a single FL round, considering both system and statistical heterogeneity. In resource-constrained wireless FL, variable network connectivity results in different time costs for uploading local gradients, while the non-IID characteristics of local datasets cause client gradients to contribute differently to FL convergence \cite{salehi2021federated}. Furthermore, fluctuating network conditions require real-time decision-making. To capture the impact of each local gradient on the convergence of single-round FL, we analyze the one-step convergence of the FedAvg algorithm in the next section.


\section{Convergence Analysis of Federated Learning Algorithm}\label{sec:Convergence_Analysis}

In the subsequent convergence analysis, we introduce several new symbols to simplify the proof. First, let $g_{n,t}$ denote the cumulative full gradient, defined as
\begin{align}
    g_{n,t} &= \eta_\text{l} \sum_{i=1}^{E}{\nabla F_n\left(w_{n,t}^{\left(i-1\right)}\right)},
\end{align}
where 
\begin{equation}\label{equ:local_full_gradient}
    \nabla F_n\left(w_{n,t}^{\left(i-1\right)}\right)=\frac{1}{|\mathcal{D}_n|}\sum_{x\in\mathcal{D}_{n}}{\nabla \ell\left(w_{n,t}^{\left(i-1\right)};x\right)}
\end{equation}
is the local full gradient for the $i$-th local training.

Furthermore, we define the aggregated stochastic gradient, $\tilde{g}_{t}$, the ideally-uploaded stochastic gradient, $\tilde{\bar{g}}_{t}$, and the ideally-uploaded full gradient, $\bar{g}_{t}$, given by
\vspace{-0.15cm}
\begin{align}
    \tilde{g}_{t}&=\frac{1}{\left|\mathcal{N}_t\right|}\sum_{n\in \mathcal{N}_t}{\tilde{g}}_{n,t},\\
    \tilde{\bar{g}}_{t} &= \frac{1}{N}\sum_{n=1}^{N}\tilde{g}_{n,t},\\
    \bar{g}_{t} &= \frac{1}{N}\sum_{n=1}^{N}{g}_{n,t}.
\end{align}
\vspace{-1.3cm}

\subsection{Assumptions}
The following assumptions are made regarding the properties of the objectives for both the clients and the server:

\begin{assumption}[$L$-Smoothness]\label{assumption:L-Smoothness}
	Let $\|\cdot\| $ denote the $\ell_2$-norm of a vector. The local objective $F_n(\cdot)$ for all $n \in \mathcal{N}$ is $L$-smooth, inducing that for all $w_1$ and $w_2$, it holds
	\begin{equation}
		\|\nabla F_n(w_1) - \nabla F_n(w_2)\| \leq L \|w_1 - w_2\|.
	\end{equation}
\end{assumption}

\begin{assumption}[Unbiased Sampling and Bounded Local Variance]\label{assumption:Unbiasedness and Bounded Local Variance}
	For all $n \in \mathcal{N}$ and for a mini-batch $\mathcal{B}_{n,t,i-1}$ uniformly and randomly sampled from $\mathcal{D}_n$, the local stochastic gradient $\widetilde{\nabla}F_n(w)$ is unbiased and has bounded variance with respect to the local full gradient $\nabla F_n(w)$, i.e.,
    \vspace{-0.15cm}
	\begin{equation}
		\mathbb{E}_{\mathcal{B}_{n,t,i-1}}\left[\widetilde{\nabla}F_n(w)\right] = \nabla F_n(w),
	\end{equation}
    and
    \vspace{-0.15cm}
	\begin{equation}
		\mathbb{E}_{\mathcal{B}_{n,t,i-1}}\left[\left\|\widetilde{\nabla}F_n(w) - \nabla F_n(w)\right\|^2\right] \leq \sigma_\text{l}^2.
	\end{equation}
    \vspace{-0.6cm}
\end{assumption}

\begin{assumption}[Bounded Global Variance]\label{assumption:Bounded Global Variance}
    The global variance, defined as the mean squared difference between the local full gradient $\nabla F_n(w)$ and the global full gradient $\nabla F(w)$, is bounded, i.e.,
    \vspace{-0.2cm}
	\begin{equation}
		\frac{1}{N}\sum_{n=1}^N\left\|\nabla F_n(w) - \nabla F(w)\right\|^2 \leq \sigma_\text{g}^2.
	\end{equation}
    \vspace{-0.6cm}
\end{assumption}

The value of global variance $\sigma_\text{g}^2$ captures the degree of non-IID nature of the local data distributions. Specifically, a larger $\sigma_\text{g}^2$ indicates a greater divergence among local gradients, reflecting higher statistical heterogeneity across clients; conversely, a smaller $\sigma_\text{g}^2$ suggests less divergence and lower heterogeneity. 
It is worth noting that convergence analyses often assume a bounded squared norm of the local stochastic gradient $\widetilde{\nabla}F_n(w)$ in \cite{Li2020On, salehi2021federated}, which should be satisfied by gradient shearing technology. However, we adopt an alternative one, i.e., Assumption \ref{assumption:Bounded Global Variance}, with fewer restrictions, which broadens the scope of applicability and providing stronger convergence guarantees under more general conditions.

 

\subsection{One-Step Convergence of FedAvg Algorithm under non-IID Settings}

To facilitate our convergence analysis, we present the following key lemmas, whose detailed proof provided in Appendix \ref{proof: Lemma 1} and \ref{proof: Lemma 2}.

\begin{lemma}[Bounded Local Drift]\label{lemma:Local Drift}
    Let Assumption \ref{assumption:Unbiasedness and Bounded Local Variance} hold and the local learning rate satisfy $\eta_\text{l} \leq \frac{1}{\sqrt{8}EL}$. Then the local drift is bounded, given by
    \begin{equation}
    \begin{aligned}
        \mathbb{E}\left\| w_{n,t}^{(E)} - w_{t}\right\|^2  &= \mathbb{E}\left\| \tilde{g}_{n,t} \right\|^2\\
        &\leq 12 \eta_\text{l}^2 E^2\left\| F_n(w_{t})\right\|^2 + 3\eta_\text{l}^2 E \sigma_\text{l}^2.
    \end{aligned}  
    \end{equation}
\end{lemma}

\begin{lemma}[Bounded Average Squared Norm]\label{lemma:Bounded Average Squared Norm}
    Let Assumption \ref{assumption:Bounded Global Variance} hold. Then the average squared norm of the local full gradient is bounded, given by
    \begin{equation}
        \frac{1}{N}\sum_{n=1}^N\left\|\nabla F_n(w)\right\|^2 \leq \left\|\nabla F(w)\right\|^2 + \sigma_\text{g}^2.
    \end{equation}
\end{lemma}

Leveraging Lemma \ref{lemma:Local Drift} and Lemma \ref{lemma:Bounded Average Squared Norm}, we derive the one-step convergence bound for the FedAvg algorithm under non-IID settings, as stated in the following theorem. The proof is detailed in Appendix \ref{proof: Theorem 1}.

\begin{theorem}[One-Step Convergence of FedAvg Algorithm]\label{theorem:One-Step Convergence}
	Let Assumptions 1-3 hold, and the local learning rate satisfy $\eta_\text{l} \leq \frac{1}{\sqrt{8}EL}$. Given the set $\mathcal{N}_t$ of clients which successfully upload gradients in the $t$-th FL round, the one-step convergence bound of the FedAvg algorithm is given by
	\begin{equation}\label{equ:One-Step_Convergence}
		\mathbb{E}[F(w_{t+1})] - F(w_{t}) \leq C_1 \|\nabla F(w_{t})\|^2 + C_2 \mathbb{E} \|\tilde{\bar{g}}_{t}-\tilde{g}_{t}\|^2 + C_3,
    \end{equation}
    where
    \vspace{-0.2cm}
    \begin{equation*}
	   \begin{aligned}
		  C_1 &= 2 \eta_\text{g} (1 - \eta_\text{l} E)^2  - \frac{1}{2} \eta_\text{g} + 12 \eta_\text{g}^2 \eta_\text{l}^2 E^2 L+ 24 \eta_\text{g} \eta_\text{l}^4 E^4 L^2, \\
		  C_2 &= \eta_\text{g} (1 + \eta_\text{g} L),\\
		  C_3 &= \left[2 \eta_\text{g} (1 - \eta_\text{l} E)^2  + 12 \eta_\text{g}^2 \eta_\text{l}^2 E^2 L + 24 \eta_\text{g} \eta_\text{l}^4 E^4 L^2\right] \sigma_\text{g}^2 \\
            &+ \left(\frac{1}{2N} \eta_\text{g} \eta_\text{l}^2 E^3 + 3 \eta_\text{g}^2 \eta_\text{l}^2 E L + 6 \eta_\text{g} \eta_\text{l}^4 E^3 L^2\right)\sigma_\text{l}^2.
	   \end{aligned}
    \end{equation*}
\end{theorem}


According to Theorem \ref{theorem:One-Step Convergence}, the convergence bound of the FedAvg algorithm depends on three terms: 1) the squared norm of the previous global full gradient, 2) the estimated squared bias between the ideally-uploaded stochastic gradient $\tilde{\bar{g}}_{t}$ and the aggregated stochastic gradient $\tilde{g}_{t}$, and 3) a constant determined by the FL settings. Among these, the second term is particularly significant in practice because it is governed by the upload results in the current round, while the first and third terms are determined only by the initialization and FL hyperparameters. Therefore, our strategy for spectrum and power allocation is designed to minimize the second term, thereby improving the convergence of FL.

To achieve this minimization, we focus on controlling the aggregated stochastic gradient $\tilde{g}_{t}$ via the upload success indicator $s_{n,t}$ defined in \eqref{equ:successful_upload_gradient}. In this paper, the upload success indicator $s_{n,t}$ requires sequential decision-making across multiple coherent time slots. This requirement exponentially increases the decision space for traditional methods, rendering them inefficient. To address this challenge, we propose an intelligent spectrum selection and power allocation strategy based on the RL framework. This strategy directly accounts for the combined effects of these two steps, as detailed in the next section.

\section{MARL-based Resource Allocation Strategy}\label{sec:MARL}

In this section, we formulate the resource allocation problem for wireless FL as a Dec-POMDP, design its components, and propose a QMIX-based solution.

\subsection{Dec-POMDP Formulation}

To mitigate the dimensionality explosion in the single-agent setting, we formulate the gradient upload problem for FL clients under spectrum-sharing schemes as an MARL problem. Specifically, we model the local gradient uploading process within a single FL round as an episode. In this setting, agents\footnote{In this paper, ``clients'' and ``agents'' are used interchangeably.} deployed at the FL clients independently allocate uplink sub-bands and transmit power in each time slot, based on local observations. Through repeated interactions with the environment, these agents progressively learn an optimal strategy for spectrum and power allocation. However, due to potential signal interference in shared sub-bands, the Markov game may exhibit competitive characteristics among agents, which can hinder learning efficiency. To overcome this challenge, we first transform the problem into a fully cooperative game by assigning a global reward for joint actions. We then perform credit assignment to further encourage cooperation, thus enhancing the FL training performance.

In this cooperative framework, each agent only has access to partial observations of the environment. Mathematically, we formulate this problem as a Dec-POMDP, represented by the tuple $\mathcal{G} = \langle \mathcal{S}, \mathcal{A}, P, r, \mathcal{Z}, \mathcal{N}, \gamma \rangle$. In this formulation, \( s \in \mathcal{S} \) represents the state of the environment. At each time step, corresponding to a time slot $t_s$, each agent \( n \in \mathcal{N} \) independently selects an action \( a_{n, t_s} \in \mathcal{A} \), forming the joint action \( \boldsymbol{a} \in \mathcal{A}^N \). This joint action induces state transitions, which are governed by the transition function \( P(s'|s, \boldsymbol{a}): \mathcal{S} \times \mathcal{A}^N \times \mathcal{S} \to [0, 1] \). Following the state transition, all agents share a global reward \( r(s, \boldsymbol{a}): \mathcal{S} \times \mathcal{A}^N \to \mathbb{R} \), and \( \gamma \in [0, 1) \) is the discount factor. Since the environment is partially observable, each agent receives a local observation \( z \in \mathcal{Z} \), determined by the observation function \( O(s, \boldsymbol{a}): \mathcal{S} \times \mathcal{A} \to \mathcal{Z} \).

To solve the Dec-POMDP, we propose to employ the QMIX algorithm, an MARL approach under the centralized training and decentralized execution (CTDE) paradigm \cite{rashid2020monotonic}. During centralized training, all agents collaboratively explore the environment. A mixing network decomposes the global reward to assign a credit to each agent, guiding them in training their individual deep Q-networks (DQNs) to learn the optimal policy. In the decentralized execution phase, each agent inputs its local observation into its trained DQN and autonomously selects the appropriate sub-bands and transmit power at each time step. The key components of the QMIX-based resource allocation strategy are detailed below.



\subsection{Observation and State Space}

To ensure user privacy in the decentralized execution phase, we dictate that each agent can only access partial local observations to infer the global state. Specifically, the local observation of agent $n$ at time slot $t_s$ of $t$-th FL round, denoted as $ z_{n,t,t_s}\in\mathcal{Z}$, includes the following components:
\begin{itemize}
    \item \textbf{Large-Scale Fading Information:} This component captures the large-scale fading $\alpha_{n,t}$ from client $n$ to the BS, which reflects the channel conditions. It also includes the large-scale fading $\alpha_{n',t}$ from other clients to the BS, which can be obtained from the BS that broadcasts such information after collecting it from all participating clients.


    \item \textbf{Small-Scale Fading Information:} This component includes the small-scale fading $h_{n,t,t_s}^{(c)}, \forall c\in \mathcal{C}$ from client $n$ to the BS across all sub-bands. This information reflects the dynamic, short-term variations in the channel conditions that directly impact the instantaneous channel quality.

    \item \textbf{Transmission Progress Information:} This component includes two key elements: 1) the remaining transmission volume for client $n$, $ \Delta_{n,t,t_s} = S - \sum_{\tau_s=1}^{t_s} C_{n,t,\tau_s} \times T_d $, and 2) the number of time slots left in the current round, $l_{t_s}= T_s - t_s $. Together, these variables capture the urgency of the transmission, reflecting both the remaining data to be transmitted and the time constraints imposed by the current round.
    
    \item \textbf{Gradient Information:} Deep neural networks in FL typically involve  millions of parameters, making it challenging to directly incorporate model gradients into the observation. To address this, we choose the gradient deviation, defined as $\delta^2_{n,t} = \| \tilde{\bar{g}}_{t} - \tilde{g}_{n,t} \|^2$, as the observation for each client, inspired by the result of convergence analysis. However, due to the distributed nature of FL, directly obtaining $\tilde{\bar{g}}_{t}$ is infeasible, as it requires aggregating the cumulative stochastic gradient $\tilde{g}_{n,t}$ from all clients. To address this, we approximate $\tilde{\bar{g}}_{t}$ using the aggregated stochastic gradient $\tilde{g}_{t-1}$ from the previous FL round. Specifically, for client $n$, $\tilde{g}_{t-1}$ can be computed as the difference between the global models $w_t$ and $w_{t-1}$, which has been broadcast by the BS in $t$-th and $(t\!-\!1)$-th FL rounds, respectively. Therefore, this approximation reduces communication overhead by leveraging client-side storage. As a result, the gradient information for the current FL round is then expressed as the estimated gradient deviation $ \tilde{\delta}^2_{n,t} = \| \tilde{g}_{t-1} - \tilde{g}_{n,t} \|^2$, which reflects the property of local gradient of client $n$.
    

    \item \textbf{FL Round Information:} As FL training converges, the magnitude of the gradients to be uploaded gradually decreases \cite{guo2020analog}. Recall that the local gradient uploading process within a single FL round is treated as an episode. However, as the number of episodes increases during RL training, the environment continues to evolve, presenting significant challenges due to its non-stationary nature. Moreover, the use of experience replay in RL training exacerbates this non-stationarity, as the random sampling from replay memory fails to capture the underlying dynamics of the environment. To address this issue, inspired by \cite{foerster2017stabilising}, we incorporate the current FL round $t$ as a ``fingerprint'' within the observation and state design, allowing us to record the current training phase and providing additional context for long-term strategy optimization.
\end{itemize}

Since the RL episode only corresponds to one FL round, we omit the FL round index $t$ for simplicity in the remainder of this paper whenever possible. The local observation for agent $n$ at time slot $t_s$ is thus summarized as
\begin{equation}
z_{n, t_s}=\left[\{\alpha_{n}\}_{n\in \mathcal{N}},\{h_{n,t_s}^{(c)}\}_{c\in \mathcal{C}}, \Delta_{n,t_s}, l_{t_s}, \tilde{\delta}^2_n, t\right]\!.
\end{equation}

In the centralized learning phase, the global state of the environment is fully observable. After eliminating redundancies in the local observations, the global state at time slot $t_s$ is defined as
\begin{equation}
    \begin{aligned}
    s_{t_s}=&\left[\{\alpha_{n}\}_{n\in \mathcal{N}},\{h_{n,t_s}^{(c)}\}_{n\in \mathcal{N}, c\in \mathcal{C}}, \right.\\
    &~\left.\{\Delta_{n,t_s}\}_{n\in \mathcal{N}}, l_{t_s}, \{\tilde{\delta}^2_n\}_{n\in \mathcal{N}}, t\right]\!.
    \end{aligned}
\end{equation}

\subsection{Action Space}
The action of agent $n$ at time slot $t_s$ is defined as $a_{n,t_s} \in \mathcal{A} \equiv [c_{n,t_s}, p_{n,t_s}]$, where $c_{n,t_s} \in \mathcal{C}$ denotes the sub-band selection and $p_{n,t_s} \in \mathcal{P}$ represents the transmit power allocation with $P$ discrete levels. Discretizing power control values simplifies RL training and makes practical circuit implementation easier. Consequently, the action space comprises $C\times P$ distinct actions. The individual actions $a_{n,t_s}$ of all agents collectively form a joint action, denoted as $\boldsymbol{a}_{t_s}\in \mathcal{A}^N$. 



\subsection{Reward Design}



After taking actions, all agents receive a global reward $r_{t_s}$. The convergence analysis in Section \ref{sec:Convergence_Analysis} indicates that FL performance can be improved by minimizing the second term in \eqref{equ:One-Step_Convergence}, which can be expanded as
\begin{equation}
    \begin{aligned}
    \left\|\tilde{\bar{g}}_{t} -\tilde{g}_{t}\right\|^2 &= \left\|\tilde{\bar{g}}_{t} -\frac{1}{\left|\mathcal{N}_t\right|}\sum_{n\in \mathcal{N}_t}{\tilde{g}}_{n,t}\right\|^2\\
    &= \frac{1}{\left|\mathcal{N}_t\right|^2}\left\|\sum_{n\in \mathcal{N}_t}{(\tilde{\bar{g}}_{t} -{\tilde{g}}_{n,t}})\right\|^2\\
    &\approx\frac{1}{\left|\mathcal{N}_t\right|^2}\left[\sum_{n\in \mathcal{N}_t}\sum_{n^{\prime}\in \mathcal{N}_t}{\langle\tilde{\delta}_{n,t},\tilde{\delta}_{n^{\prime},t}\rangle}\right]\!,
\end{aligned}
\label{equ:bound_second_term}
\end{equation}
where $\tilde{\bar{g}}_{t}$ is approximated by $\tilde{g}_{t-1}$. 
However, calculating \eqref{equ:bound_second_term} requires determining $\mathcal{N}_t$, which can only be done at the end of the episode. Therefore, directly taking the negative of \eqref{equ:bound_second_term} as the reward means that the agent can only receive a non-zero reward at the end of each episode, which makes it difficult to fully guide these sequential decisions.
To enable more effective learning and adaptation in the resource allocation process, it is necessary to design the reward to be decomposed at each time step. We observe that \eqref{equ:bound_second_term} encourages us to upload more client gradients and select appropriate clients for better approximation of the ideal case. To facilitate decomposition, we heuristically assign the convergence-related reward for the $t$-th FL round as
\begin{equation}
r^{\text{c}}_{t}=\lambda_1\left|\mathcal{N}_t\right|-\lambda_2\sum_{n\in \mathcal{N}_t}{\left[{\tilde{\delta}^2_{n,t}}+\!\!\!\!\!\!\!\sum_{n^{\prime}\in \mathcal{N}_t,n^{\prime}\neq n}\!\!\!\!\!\!{\langle\tilde{\delta}_{n,t},\tilde{\delta}_{n^{\prime},t}\rangle}\right]}\!,
\end{equation}
where the hyperparameters $\lambda_1, \lambda_2\in[0,1]$ balance the influence of the two terms in the reward function and are selected to ensure positive reward value. Moreover, the local gradient for reward calculation is only available once the client successfully uploads. Consider that, this decomposition must ensure that rewards are independent of subsequent uploads. We then derive the convergence-related reward for the $t_s$-th time slot as
\begin{equation}
\!\!\!r^{\text{c}}_{t_s}=
\begin{cases}
\begin{aligned}
&\lambda_1\!-\!\lambda_2\sum_{n\in \mathcal{N}_{t_s}}\!\!\left[\tilde{\delta}^2_{n}\!+\!\!\sum_{n^{\prime}\in \mathcal{N}_{t_s}^-}\!\!{2\langle\tilde{\delta}_{n},\tilde{\delta}_{n^{\prime}}\rangle}\right. \\
&\left.+\sum_{n^{\prime}\in \mathcal{N}_{t_s},n^{\prime}\neq n}\!\!\!{\langle\tilde{\delta}_{n},\tilde{\delta}_{n^{\prime}}\rangle}\right]\!\!,
\end{aligned}
&\!\!\text{if}~ \mathcal{N}_{t_s} \!\!\neq \!\varnothing,\\
0,&\!\!\text{otherwise,}
\end{cases}
\end{equation}
where $\mathcal{N}_{t_s}$ and $\mathcal{N}_{t_s}^-$ represent the sets of clients that have successfully upload gredients by the $t_s$-th time slot and those that have uploaded before the $t_s$-th time slot, respectively, in the current FL round. 
However, this design still fails to provide sufficient guidance to the agent. The agents receive non-zero rewards only when the clients succeed in uploading their gradients, which leads to aimless exploration in the early stage of training. To address this, we incorporate prior knowledge into the reward function, i.e., a higher total transmission rate enhances client uploads. Consequently, we redefine the reward at each time slot $t_s$ as
\begin{equation}
    r_{t_s} = \lambda_\text{c} r^{\text{c}}_{t_s} + \lambda_\text{t} \sum_{n=1}^N C_{n,t_s},
\end{equation}
where $\lambda_\text{c}$ and $\lambda_\text{t}$ are positive weights that balance and normalize the convergence- and transmission-related rewards.



\subsection{Learning Algorithm}

\begin{figure}[tbp]
    \centering
    \includegraphics[width=0.48\textwidth]{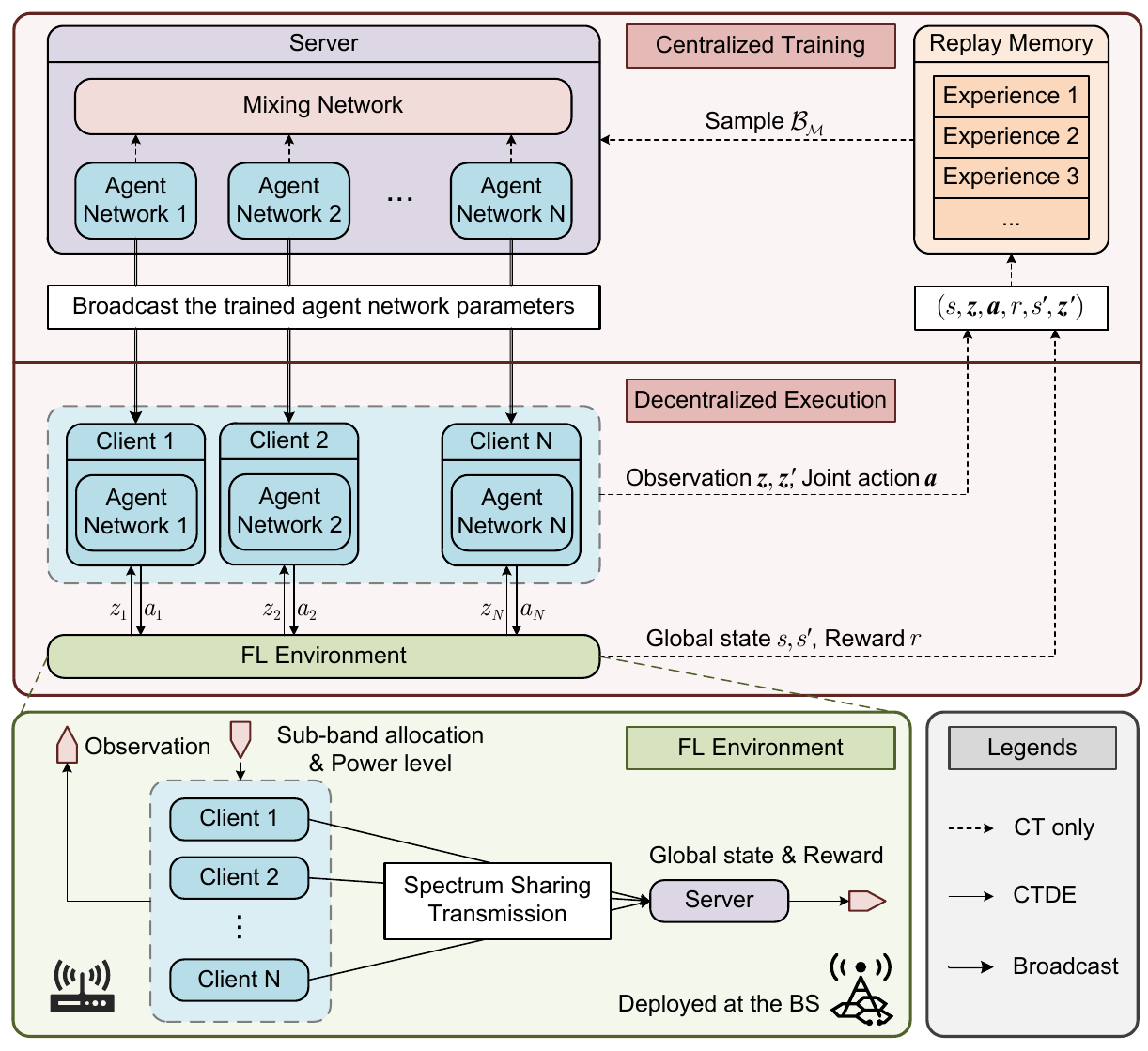}
    \vspace{-3mm}
    \caption{Workflow of the proposed resource allocation strategy: In the centralized training phase, each network is deployed at the BS, interacts with the FL environment, and learns from stored experiences. After training, the BS sends the trained agent network parameters to the corresponding clients. In the decentralized execution phase, each client independently decides its sub-band allocation and power level based on local observations.}
    \label{fig:FL_RL_workflow}
\end{figure}

Building on the previously introduced Dec-POMDP formulation, we employ the QMIX algorithm to derive the optimal spectrum and power allocation policy. As shown in Fig. \ref{fig:FL_RL_workflow}, this approach effectively coordinates multiple agents to optimize a collaborative objective with the CTDE paradigm.

\textbf{1) Training Procedure:} The centralized training is conducted at the BS side, utilizing experiences that encompass joint observations, joint actions, global states, and global rewards. These experiences are stored in a replay memory as tuples in the form $(s, \boldsymbol{z}, \boldsymbol{a}, r, s^\prime, \boldsymbol{z}^\prime)$, where the superscript $\prime$ indicates the next time slot. This centralized methodology enables the BS to leverage global information, thereby enhancing the training effectiveness of the agents.

For each agent $n$, the individual action-value function, or individual Q-value, is defined as the expected cumulative discounted return from executing action $a_n$ given observation $z_n$ under the policy $\pi_n$, given by
\begin{equation}
Q_n^{\pi_n}(z_{n}, a_{n}) = \mathbb{E}_{\pi_n}\!\!\left[ \sum_{\tau_s=t_s}^{T_s} \gamma^{\tau_s-t_s}r_{n,\tau_s}\!\mid\!  z_{n,t_s}\!=z_{n}, a_{n,t_s}\!= a_{n}\right]\!.
\end{equation}

DQNs are utilized to approximate the optimal individual Q-value $Q^*_n(z_n, a_n)=\text{max}_{\pi_n}Q^{\pi_n}_n(z_n, a_n)$ through deep neural networks, i.e., $Q_n(z_n, a_n;\theta_{\text{a},n})\approx Q^*_n(z_n, a_n)$. In a multi-agent environment, this framework is extended by defining the joint Q-value as $Q_{\text{tot}}$, which captures the value of a specific joint observation-action pair for the entire system.

\begin{figure}[tbp]
    \centering
    \vspace{-3mm}
    \includegraphics[width=0.48\textwidth]{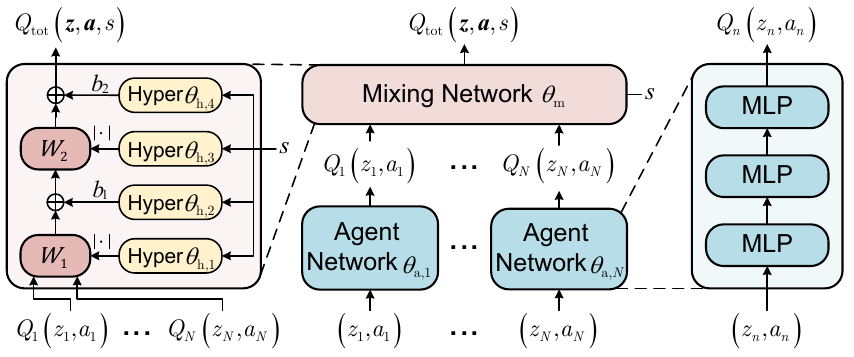}
    \caption{QMIX network: The blue block on the right represents the agent network, while the red block on the left depicts the mixing network. Within this network, the yellow block illustrates the hypernetwork, which generates the weights and biases for its layers.}
    \label{fig:qmix_network}
\end{figure}

QMIX approximates the optimal joint Q-value $Q^*_{\text{tot}}(\boldsymbol{z}, \boldsymbol{a}, s)$ using a set of networks parameterized by $\boldsymbol{\theta}=\{\boldsymbol{\theta}_\text{a},\theta_\text{m},\boldsymbol{\theta}_\text{h}\}$, as shown in Fig. \ref{fig:qmix_network}. This parameter set includes the individual agent networks $\boldsymbol{\theta}_\text{a}= [ \theta_{\text{a},1},\dots,\theta_{\text{a},N} ]$, a mixing network $\theta_\text{m}$, and a set of hypernetworks $\boldsymbol{\theta}_\text{h}$. Each agent network $\theta_{\text{a},n}$ processes its own observation-action pair to produce $Q_n(z_n, a_n;\theta_{\text{a},n})$. The mixing network then combines the outputs of all agent networks to produce the joint Q-value $Q_{\text{tot}}$. Additionally, the hypernetworks dynamically generate the weights and biases for the mixing network $\theta_\text{m}$ based on the global state $s$, ensuring that the joint Q-value function $Q_{\text{tot}}$ maintains a monotonic relationship with individual Q-values $Q_n$. In summary, QMIX integrates all these networks to compute $Q_{\text{tot}}(\boldsymbol{z}, \boldsymbol{a}, s; \boldsymbol{\theta})$ as
\begin{equation}
\begin{aligned}
    \!\!Q_{\text{tot}}(\boldsymbol{z},\boldsymbol{a}, s; \boldsymbol{\theta})\! =\! \text{Mixing}(&Q_1(z_1, a_1;\theta_{\text{a},1}),\dots, \\&Q_N(z_N, a_N;\theta_{\text{a},N}); \theta_\text{m}(s;\boldsymbol{\theta}_\text{h})).
\end{aligned}
\end{equation}

To mitigate the over-estimation bias inherent in DQN, we employ the double deep Q-Network (DDQN) approach in the training process. This improvement leads to more accurate Q-value estimations and stabilizes the training process. Consequently, the QMIX network parameters $\boldsymbol{\theta}$ are optimized by minimizing the following loss function
\begin{equation}\label{equ:QMIX_objective}
    \mathcal{L}(\boldsymbol{\theta})=\sum_{\mathcal{B}_{\mathcal{M}}}\left[\left(y^{\text{tot}}- Q_{\text{tot}}(\boldsymbol{z},\boldsymbol{a}, s; \boldsymbol{\theta})\right)^2\right],
\end{equation}
where $\mathcal{B}_{\mathcal{M}}$ represents a batch of experiences uniformly sampled from the replay memory $\mathcal{M}$, and the target Q-value is computed as
\begin{equation}
    y^{\text{tot}} = r + \gamma Q_{\text{tot}}(\boldsymbol{z}^\prime,\boldsymbol{a}^\prime, s^\prime; \boldsymbol{\theta}^-),
\end{equation}
where the target action $\boldsymbol{a}^\prime = [ {a}^\prime_1,\dots,{a}^\prime_N]$ is determined by the online agent networks as $ {a}^\prime_n =  \arg\max_{a^\prime} Q_{n}(z^\prime_{n},a^\prime_{n}; \theta^-_{\text{a},n})$, and $\boldsymbol{\theta}^-\!$ denotes the parameter set of the target networks. These target networks are periodically updated by copying from $\boldsymbol{\theta}$ and remain fixed for several iterations to enhance stability.


Finally, to balance exploration and exploitation during training, an $\epsilon$-greedy policy is employed. In this approach, each agent selects a random action with probability $\epsilon$, encouraging exploration of the action space, and with probability $1-\epsilon$, it selects the action with the highest estimated individual Q-value, exploiting the current knowledge to maximize the expected return. The pseudo-code of the training procedure is summarized in Algorithm \ref{alg:QMIX_Spectrum_and_Power_Allocation}.

\begin{algorithm2e}[t]
\caption{QMIX-based Resource Allocation for FL Training}\label{alg:QMIX_Spectrum_and_Power_Allocation}
\SetAlgoLined

Initialize $\boldsymbol{\theta}$ for the mixing network, agent networks and\\ the hypernetwork, and $w_0$ for the FL model.

\For{each FL round $t$}{
    The BS broadcasts $w_t$ to all $N$ clients\;
    Each client $n$ computes $\tilde{g}_{n,t}$ with local dataset $\mathcal{D}_n$\;
    Reset $\Delta_{n,0}\!= S$ for all clients\;
    \For{each time slot $t_s$}{
        Update channel fading $\alpha_{n}\!$ and $h_{n,t_s}^{(c)}$\;
        \For{each agent $n$}{
            Observe $z_{n,t_s}$\;
            Choose action $a_{n,t_s}\!$ using $\epsilon$-greedy policy\;
            \If{agent $n$ finishes transmission}{
               Disable the transmission of agent $n$\;
            }
            Perform transmission and update $\Delta_{n,t_s}$\;
        }
        Get global reward $r = r_{t_s+1}$ and next state \\$s^\prime = s_{t_s+1}$\;
        Store $(s,\boldsymbol{z}, \boldsymbol{a},r,s^\prime,\boldsymbol{z}^\prime)$ in the replay memory \\$\mathcal{M}$\;
        Sample experiences $\mathcal{B}_{\mathcal{M}}$ from $\mathcal{M}$ uniformly\;
        Minimize loss in \eqref{equ:QMIX_objective} to optimize $\boldsymbol{\theta}$ using \\mini-batch SGD\;
        \If{update-interval steps have passed}{
           $\boldsymbol{\theta}^- = \boldsymbol{\theta}$\;
        }
    }
    Determine the set $\mathcal{N}_t$ of clients which successfully\\ upload gradient\;
    Update the global FL model according to \eqref{equ:global_aggregation}\;
}
\end{algorithm2e}

\textbf{2) Execution Procedure:} After training, the BS broadcasts the trained agent network parameters to the corresponding clients. During the decentralized execution phase, at each time slot $t_s$, each client updates the fading and transmission information to construct its local observation $z_{n,t_s}$. It then selects an action $a_{n,t_s}$ that maximizes its individual Q-value according to its trained agent network $\theta_{a,n}$. Afterwards, all clients proceed to transmit with the sub-band allocations and power levels determined by their selected actions. This separation of training and execution phases ensures that the training process leverages centralized information and coordination, while the execution phase maintains scalability and operational efficiency.

\section{Simulation Results}\label{sec:Simulation_Results}
In this section, we present simulation results to validate the theorem and reward design, followed by the proposed QMIX-based resource allocation strategy for wireless FL.

\subsection{Simulation Setup}

We conduct experiments on the CIFAR-10 dataset \cite{krizhevsky2009learning} using a CNN model consisting of three convolutional layers, each with a 
$3 \times 3$ kernel. Each convolutional layer is followed by batch normalization, rectified linear unit (ReLU) activation, and $2 \times 2$ max pooling. The output channels of the layers are 32, 64, and 128, respectively.

We configure a wireless FL system with $N = 10$ local clients. The FL server is stationed at a BS fixed at the center of a hexagonal cell with a side length of 500 meters, and the clients are uniformly distributed around the BS. The action space consists of sub-band selection $c_{n,t_s}$ and transmit power $p_{n,t_s}$. We set the number of sub-bands $C$ to 4, and the transmission power $p_{n,t_s} \in \{23, 10, 5, -100\}~\text{dBm}$. Specifically, when $p_{n,t_s} = -100~\mathrm{dBm}$, agent $n$ cancels the uploading during time slot $t_s$, and this setting also applies once the transmission is completed. Moreover, we use FL test accuracy versus the number of communication rounds as performance metric, as communication latency is the primary bottleneck. The primary simulation parameters of the FL task are detailed in Table \ref{tab:FL_parameters}. 

To simulate system heterogeneity, we model communication channels with time-varying multipath fading from QuaDRiGa \cite{jaeckel2014quadriga}. Specifically, we selected the Urban Microcell environment under non-line-of-sight conditions. In this environment, obstacles like buildings and walls cause significant multipath effects. We study small-scale fading by adjusting the number of clusters $n_c\in\{14, 21, 28\}$. More clusters leads to greater diversity in the signal propagation path, causing more severe fading fluctuations. 

\begin{table}[tbp] 
\centering
\caption{Simulation Parameters of the FL task}
\begin{tabular}{l|c}
\Xhline{0.8pt}
\textbf{Parameters} & \textbf{Values}  \\ 
\Xhline{0.4pt}
Total communication rounds $T$ & 100\\ 
Local epochs $E$& 3\\ 
Local learning rate $\eta_\mathrm{l}$ & 0.01\\ 
Global learning rate $\eta_\mathrm{g}$ & 1\\ 
Local batch size $B_\mathcal{D}$ & 50\\ 
Local gradient size $S$ &  9932960 bits\\ 
Bandwidth of each sub-band $B$ &  5 MHz\\ 
Duration of each time slot $T_d$ &  2 ms\\ 
Number of time slots within a round $T_s$ & 250\\ 
\Xhline{0.8pt}
\end{tabular}
\label{tab:FL_parameters}
\end{table}

\begin{table}[tbp] 
\centering
\caption{Hyperparameters of QMIX-based Resource Allocation}
\begin{tabular}{l|c} 
\Xhline{0.8pt}
\textbf{Parameters} & \textbf{Values}  \\ 
\Xhline{0.4pt}
Learning network update interval & 30\\ 
Target network update interval& 210\\ 
Replay buffer size & 10000\\
Training batch size & 512\\ 
Discount factor $\gamma$ & 0.95\\
Learning rate of agent network  &  $ 1\times e^{-5}$\\
Learning rate of mixing network &  $ 5\times e^{-4}$\\ 
\Xhline{0.8pt}

\end{tabular}
\label{tab:RL_parameters}
\end{table}

\begin{figure*}[tbp]
\setlength{\abovecaptionskip}{0pt}  
\setlength{\belowcaptionskip}{5pt} 
    \centering
    \begin{minipage}[b]{0.3\textwidth}
        \centering
        \includegraphics[width=\linewidth]{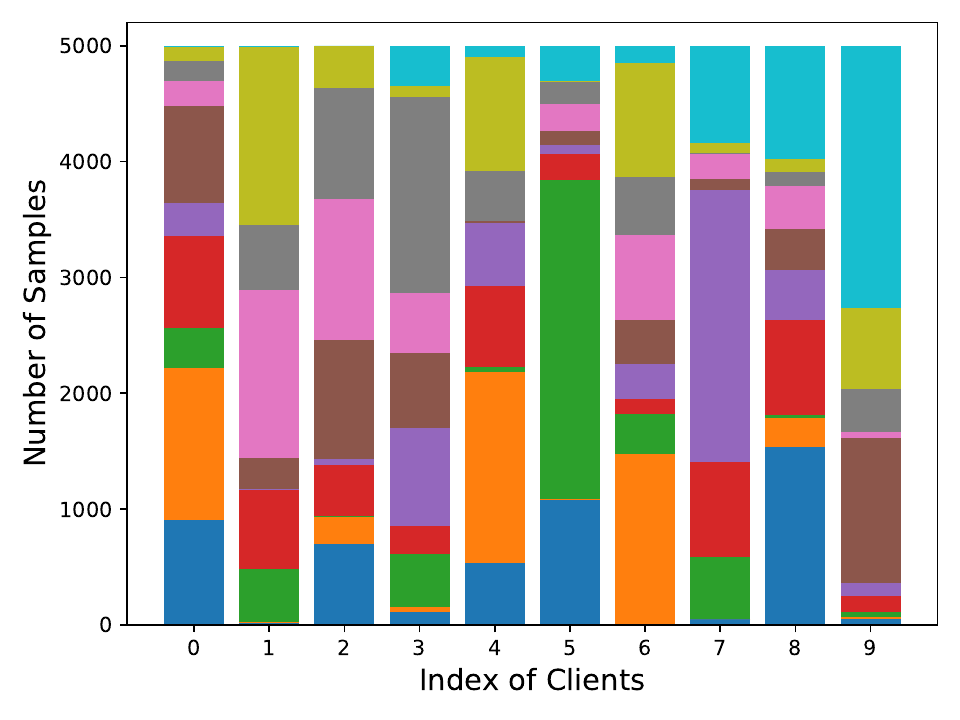}
        \centerline{a) $\alpha=0.5$}
    \end{minipage}
    \begin{minipage}[b]{0.3\textwidth}
        \centering
        \includegraphics[width=\linewidth]{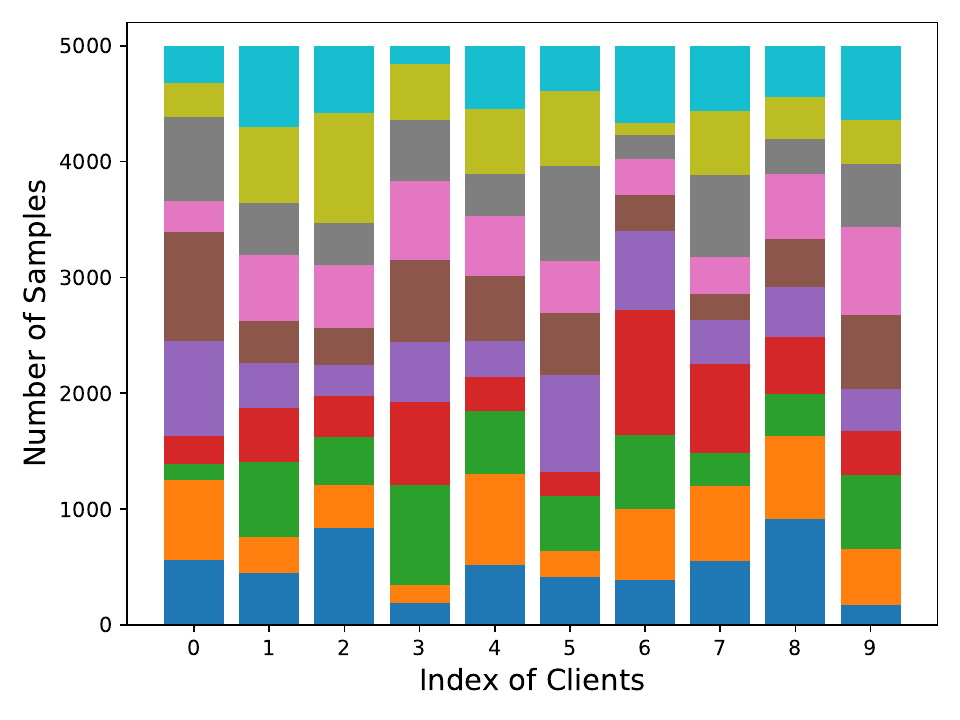}
        \centerline{b) $\alpha=5$}
    \end{minipage}
    \begin{minipage}[b]{0.3\textwidth}
        \centering
        \includegraphics[width=\linewidth]{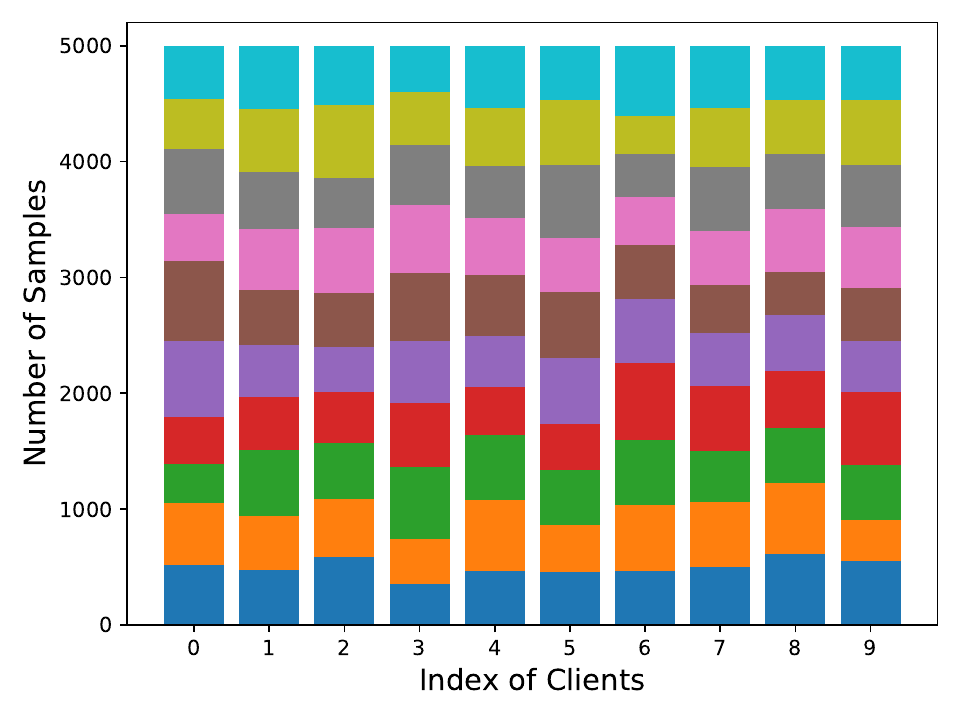}
        \centerline{c) $\alpha=50$}
    \end{minipage}
    \caption{The distribution of local training data across different degrees of statistical heterogeneity. Each column represents the dataset of an individual client, and each color indicats a distinct category of training data.
    }
    \label{fig:non-IID_partition}
\end{figure*}

To simulate statistical heterogeneity, we adopt a balanced Dirichlet partitioning approach, as established in previous studies \cite{acar2021federated,zeng2023fedlab}. In this setup, each client is assigned an equal number of data samples, and the class distribution for each client follows a Dirichlet distribution $\text{Dir}(\alpha)$. Lower values of the concentration parameter $\alpha$ result in higher levels of statistical heterogeneity. We use the non-IID partition corresponding to $\alpha \in\{ 0.5, 50, 5\}$, as shown in Fig. \ref{fig:non-IID_partition}.

The DQN of each FL client consists of three fully connected hidden layers, containing 250, 120 and 120 neurons respectively, activated by ReLU. The RMSProp optimizer \cite{ruder2016overview} is used to update the network parameters during training. We train the QMIX network for a total of 30,000 episodes and the exploration rate is linearly annealed from 1 to 0.001 over the beginning 20,000 episodes and remains constant afterwards. The hyperparameters of QMIX-based resource allocation strategy are summarized in Table \ref{tab:RL_parameters}. More specifically, each agent interacts with the recurring environment 20 times per FL round, and the global FL model is updated based on the gradient aggregated from the last interaction. Once all communication rounds of the FL process are complete, we restart the FL task and reset the channel while continue the training procedure of the RL models. In addition, we fix the concentration parameter $\alpha$ and the number of clusters $n_c$ to 0.5 and 21, respectively, during the training phase, but vary its value during the testing phase to verify the robustness of the proposed method.

\subsection{Performance Evaluation}

We compare the proposed QMIX-based resource allocation strategy against the following three heuristic baseline methods.
\begin{itemize}
    \item \textbf{Max sum rate baseline:} The global server aggregates channel information from all clients and employs a brute-force search to maximize the sum capacity across all clients for every time slot in a centralized manner. However, the exponential complexity of this algorithm renders it impractical for real-world applications.
    \item \textbf{Max individual rate baseline:} Each FL client autonomously selects its optimal sub-band with the maximum transmit power for every time slot regardless the other client in a distributed manner.
    \item \textbf{Random baseline:} Each FL client independently and randomly selects its sub-band and uses the maximum transmit power for every time slot in a distributed manner.
\end{itemize}


Additionally, we establish the upper bound of FL accuracy with the perfect communication. Specifically, we assume that all gradients in each round of FL are successfully uploaded, i.e., the second item in the convergence bound defined in Section \ref{sec:Convergence_Analysis} is equal to zero, which represents the best convergence performance.

\begin{figure*}[h]
    \centering
    \begin{minipage}[b]{0.3\textwidth}
        \centering
        \includegraphics[width=\linewidth]{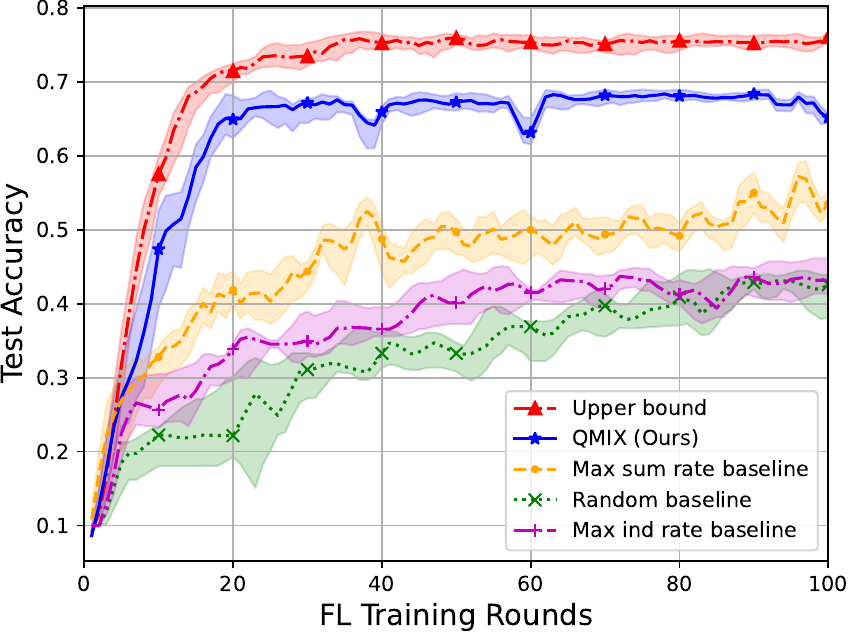}
        \centerline{a) $\alpha=0.5, n_c=14$}
    \end{minipage}
    \hspace{0.01\textwidth}
    \begin{minipage}[b]{0.3\textwidth}
        \centering
        \includegraphics[width=\linewidth]{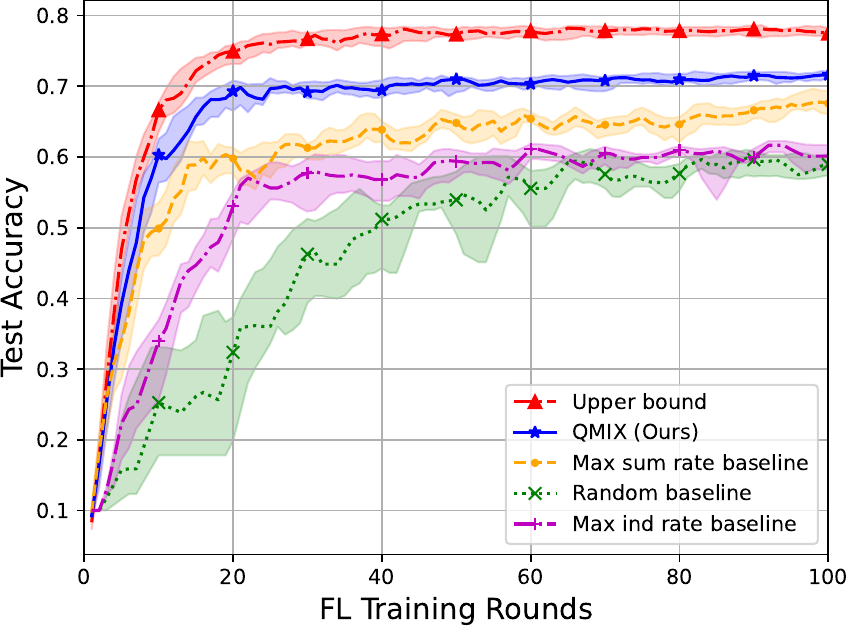}
        \centerline{b) $\alpha=5, n_c=14$}
    \end{minipage}
    \hspace{0.01\textwidth}
    \begin{minipage}[b]{0.3\textwidth}
        \centering
        \includegraphics[width=\linewidth]{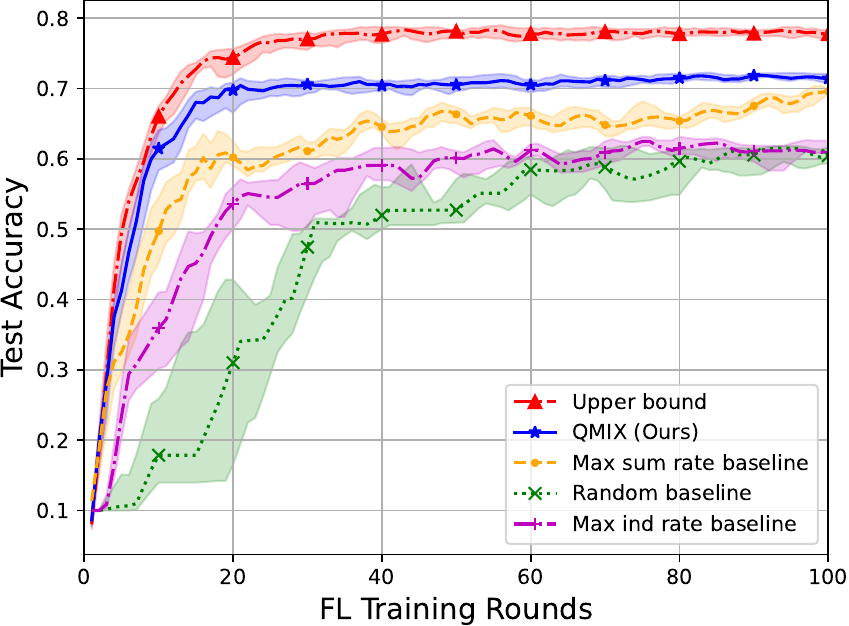}
        \centerline{c) $\alpha=50, n_c=14$}
    \end{minipage}
    \begin{minipage}[b]{0.3\textwidth}
        \centering
        \includegraphics[width=\linewidth]{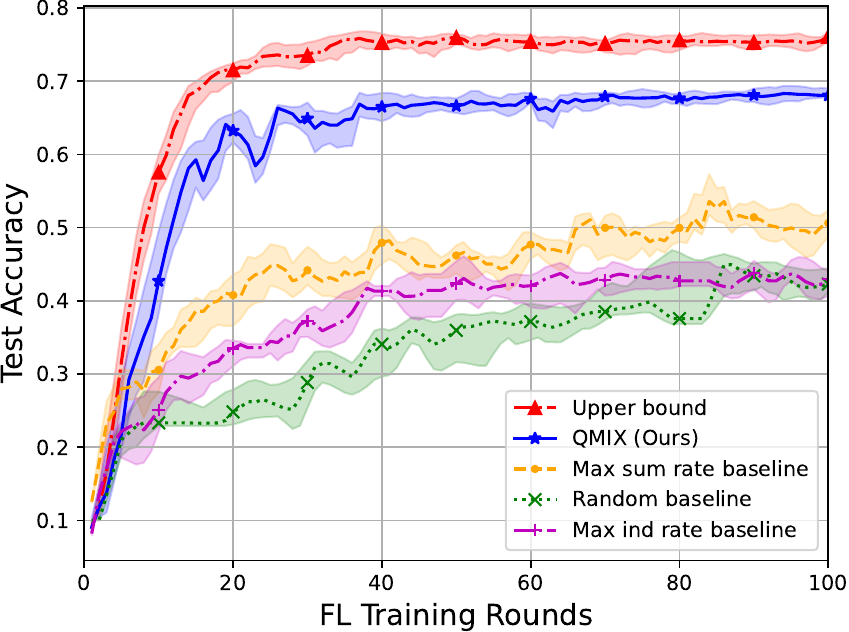}
        \centerline{d) $\alpha=0.5, n_c=21$}
    \end{minipage}
    \hspace{0.01\textwidth}
    \begin{minipage}[b]{0.3\textwidth}
        \centering
        \includegraphics[width=\linewidth]{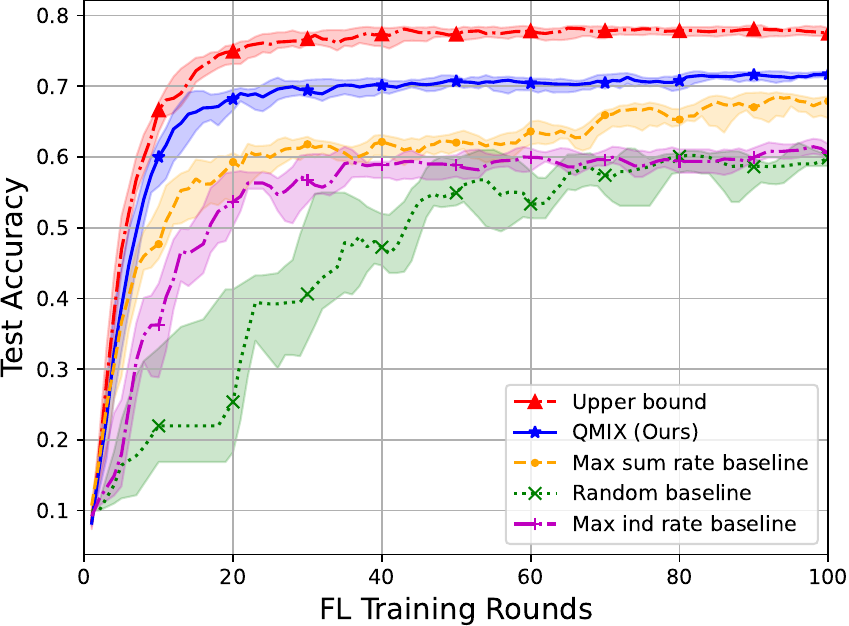}
        \centerline{e) $\alpha=5, n_c=21$}
    \end{minipage}
    \hspace{0.01\textwidth}
    \begin{minipage}[b]{0.3\textwidth}
        \centering
        \includegraphics[width=\linewidth]{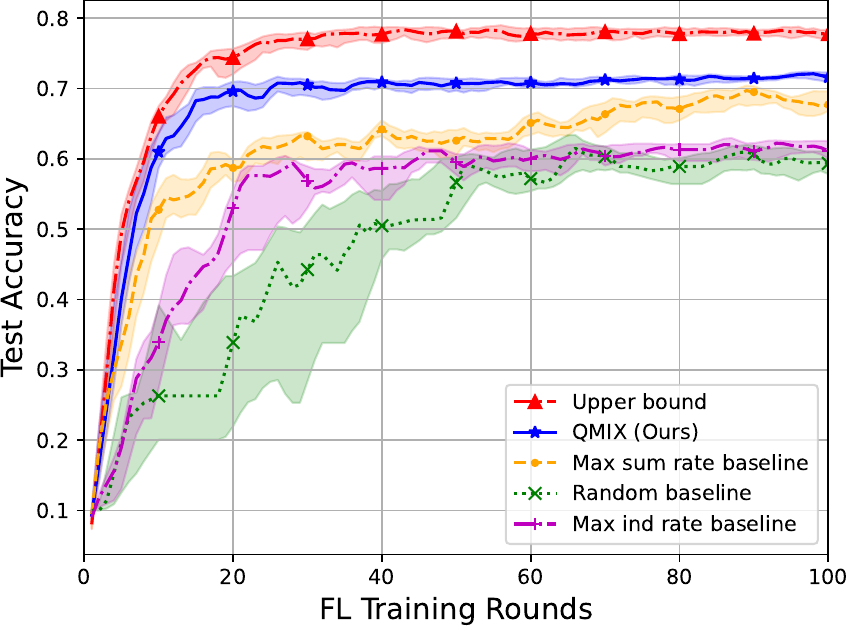}
        \centerline{f) $\alpha=50, n_c=21$}
    \end{minipage}
    \begin{minipage}[b]{0.3\textwidth}
        \centering
        \includegraphics[width=\linewidth]{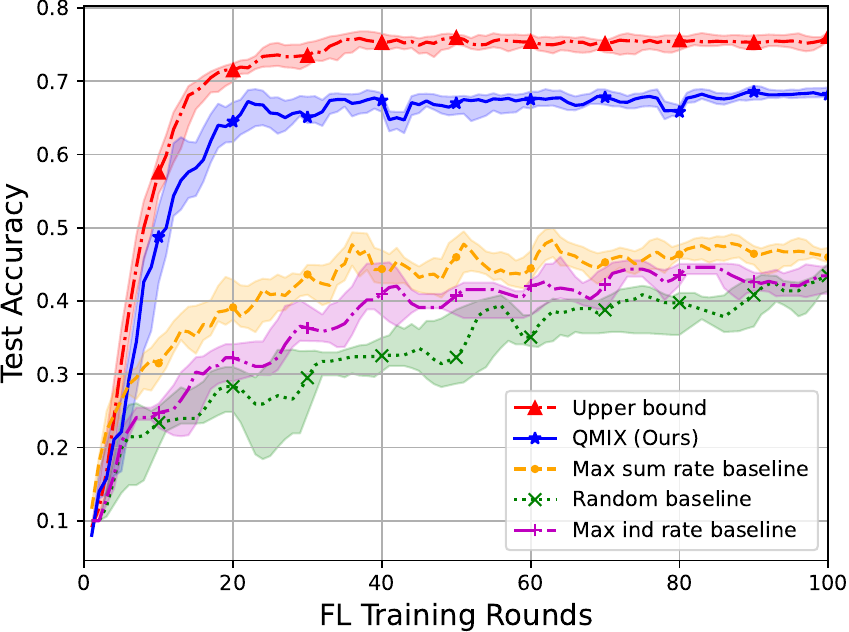}
        \centerline{g) $\alpha=0.5, n_c=28$}
    \end{minipage}
    \hspace{0.01\textwidth}
    \begin{minipage}[b]{0.3\textwidth}
        \centering
        \includegraphics[width=\linewidth]{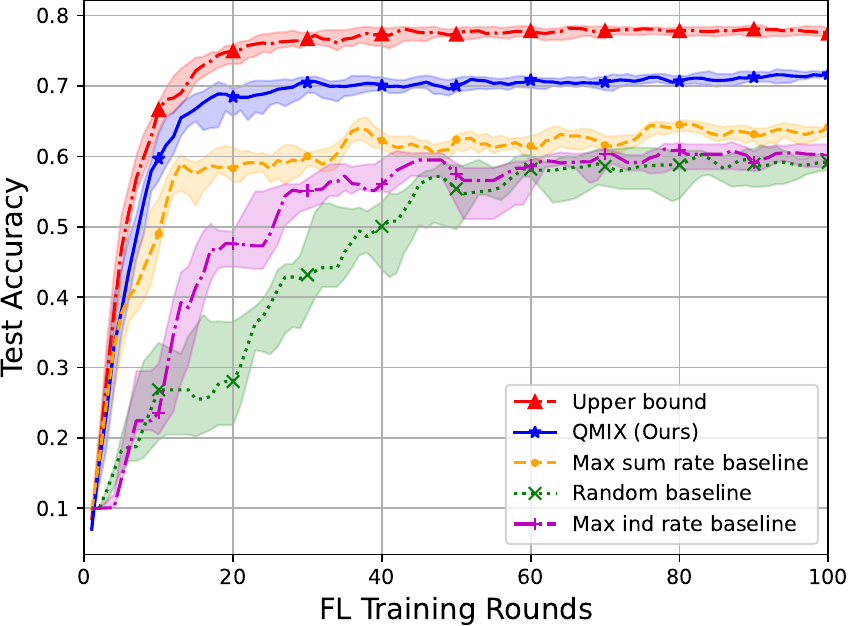}
        \centerline{h) $\alpha=5, n_c=28$}
    \end{minipage}
    \hspace{0.01\textwidth}
    \begin{minipage}[b]{0.3\textwidth}
        \centering
        \includegraphics[width=\linewidth]{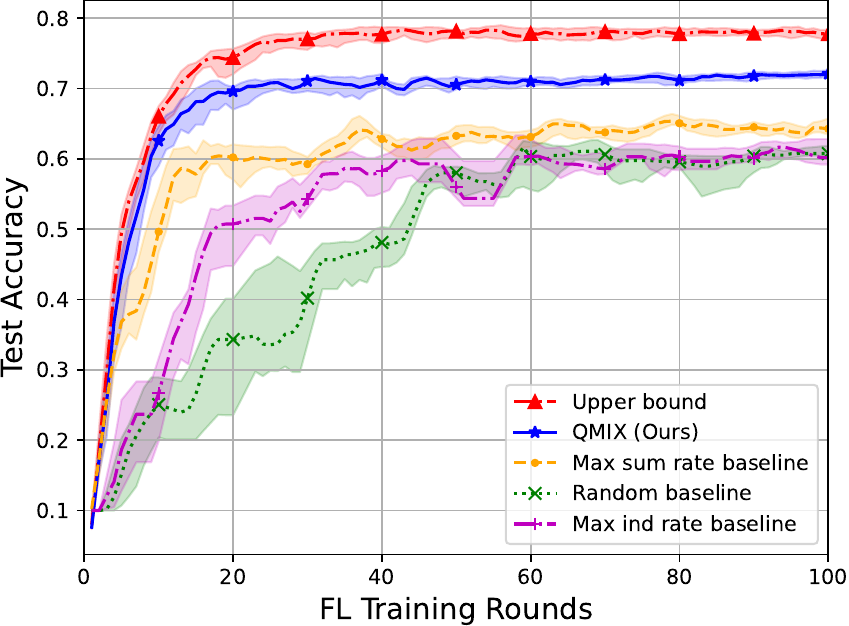}
        \centerline{i) $\alpha=50, n_c=28$}
    \end{minipage}
    \caption{Test accuracy over the number of training rounds under different levels of system and statistical heterogeneity. The concentration parameter $\alpha$ controls the statistical heterogeneity among local datasets, and the cluster number $n_c$ controls the system heterogeneity among devices.}
    \label{fig:acc_comparation}
\end{figure*}

Fig. \ref{fig:acc_comparation} shows the FL test accuracy across various resource allocation designs for each training round, with each subfigure representing a different level of system and statistical heterogeneity. The shaded area represents the 10-th to 90-th percentile range of test accuracy, estimated from 5 independent runs with randomly selected seeds. The results demonstrate that the proposed method consistently outperforms the baseline methods across all scenarios. Under high statistical heterogeneity ($\alpha = 0.5$), the method proposed in Algorithm \ref{alg:QMIX_Spectrum_and_Power_Allocation} converges to a performance level slightly below the communication-free upper bound, whereas the other baselines struggle to converge. This is because the proposed algorithm considers both system and statistical heterogeneity, enabling it to intelligently prioritize gradient uploads from clients that promote convergence. In contrast, these heuristic baselines ignore statistical heterogeneity, which impedes their convergence. Among these, the max sum rate baseline makes the most efficient use of communication resources. Although many clients experience significant transmission waste, the centralized brute-force search still achieves a relatively high transmission success rate. The max individual rate baseline attempts to efficiently utilize communication resources, but its lack of interference management negatively impacts the transmission success rate. The random baseline neither efficiently utilizes communication resources nor manages interference, resulting in the worst transmission success rate. 
As the concentration parameter $\alpha$ increases from 0.5 to 5, and further to 50, the performance of all methods improves. Moreover, as statistical heterogeneity decreases, system heterogeneity gradually dominates, and the performance of heuristic baselines that ignore statistical heterogeneity improves significantly. However, the proposed method consistently achieves the fastest convergence speed and the highest performance, demonstrating the effectiveness and robustness of this algorithm.

From the perspective of system heterogeneity, as the number of clusters $n_c$ increases, small-scale fading fluctuations become more severe, causing the performance of the max sum rate baseline to gradually decline, while the performance of other methods remains relatively stable. This is because the intense fluctuations of small-scale fading cause the main uploading client of the max sum rate baseline to continuously change, and the time-averaged performance becomes more balanced, leading to inefficient transmission. Other distributed resource allocation methods focus on the independent selection of their sub-bands, and the main uploading strategy remains relatively stable, ensuring relatively stable performance.


\begin{figure}[tbp]
    \centering
    \includegraphics[width=0.4\textwidth]{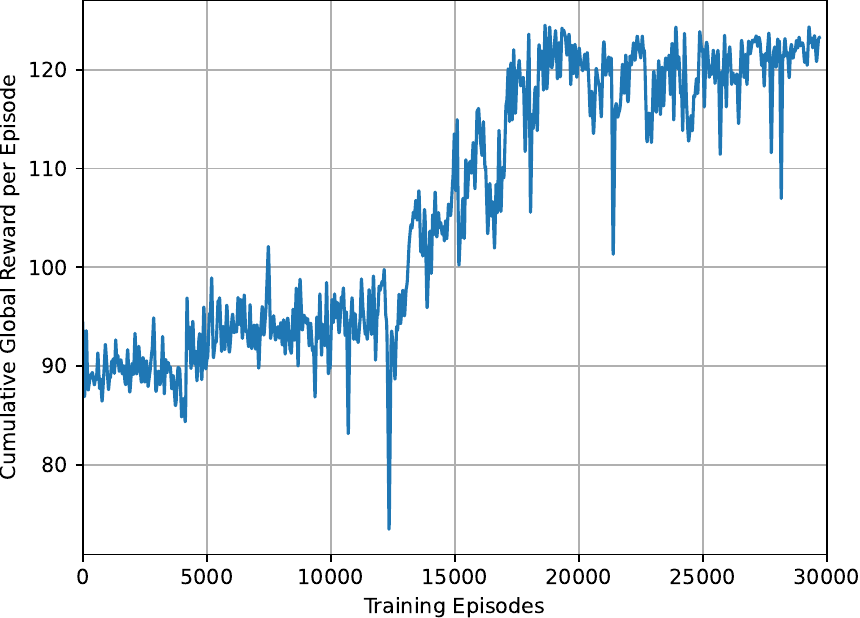}
    \vspace{-3mm}
    \caption{Cumulative global reward for each training episode with increasing training episodes. The concentration parameter and cluster number are $\alpha=0.5$ and $n_c=21$.}
    \label{fig:cumulative_global_reward}
\end{figure}

\begin{figure}[tbp]
    \centering
    \includegraphics[width=0.4\textwidth]{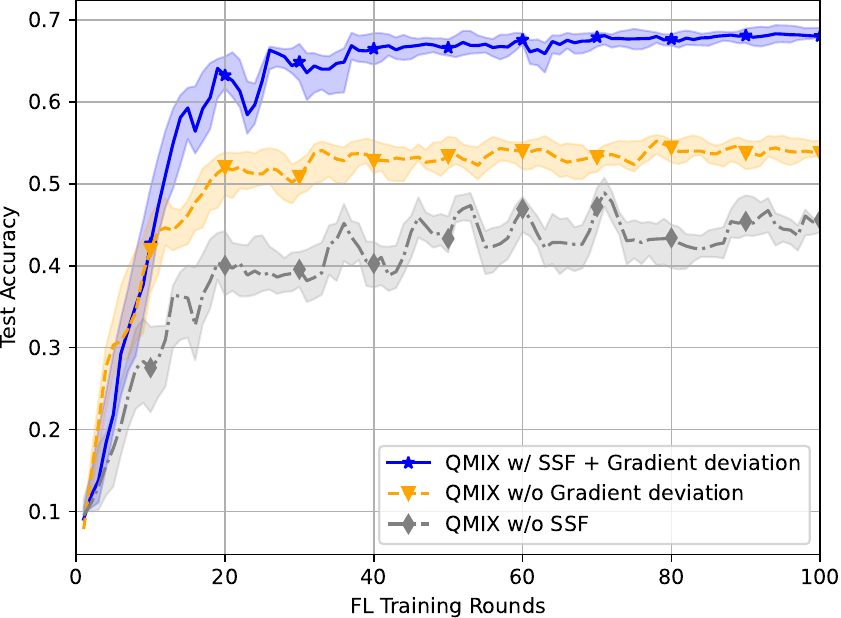}
    \vspace{-3mm}
    \caption{Test accuracy over the number of training rounds when small-scale fading or gradient derivation is unobservable. The concentration parameter and cluster number are $\alpha=0.5$ and $n_c=21$.}
    \label{fig:comparison_test}
\end{figure}

Fig. \ref{fig:cumulative_global_reward} shows the cumulative global reward of a training episode over the training episodes with the concentration parameter of $\alpha=0.5$. As shown in the figure, during the early stages of training, the model explores optimal strategies, resulting in low and highly fluctuating rewards. Midway through training, the reward value increases rapidly while the fluctuation amplitude decreases, indicating that the strategy is gradually stabilizing toward the optimal solution. In the final stages of training, despite fluctuations caused by small-scale fading and FL restarts, the cumulative global reward gradually converges. Therefore, to ensure safe convergence, we train for 30,000 episodes when evaluating the performance of the proposed strategy.

Next, we perform an ablation study to confirm the necessity of including both gradient deviation and small-scale fading. Fig. \ref{fig:comparison_test} shows the FL test accuracy for each training round when these component are not observable. To ensure a fair comparison, we maintain consistent hyperparameters and network structure and set the observation values for gradient deviation and small-scale fading to 1 during both the training and testing phases. The experiment demonstrates that failing to observe gradient deviation reduces the FL training performance, while the absence of small-scale fading observation worsens convergence and increases result instability. When clients cannot distinguish gradient deviation, agents can only identify and exploit system heterogeneity, making it difficult to explore the impact of different client gradients, leading to a relatively fixed strategy. However, due to the non-IID distribution of client data, this fixed strategy introduces model bias, thereby reducing FL training performance. In contrast, when clients lack small-scale fading observations, they will be unable to distinguish their own sub-bands. This blind sub-band selection increases the probability of strong interference during gradient uploads, reducing overall transmission efficiency. Moreover, as these clients cannot perceive favorable channel resources, the likelihood of upload failure increases, leading to instability in FL convergence.


\section{Conclusion}\label{sec:Conclusion}

In this paper, we propose a resource allocation strategy for wireless FL that accounts for the rapid fluctuations in channel conditions within FL rounds. To achieve this, we first conduct a convergence analysis to quantify the impact of local gradient uploads on FL performance. Based on this analysis, we formulate the resource allocation problem within a Dec-POMDP framework, carefully designing the observation space and reward function to facilitate efficient decision-making within a single FL round. To solve this problem, we adopt a QMIX-based learning algorithm, leveraging a CTDE paradigm to enhance the scalability and practicality of the proposed scheme. Simulation results demonstrate that the proposed resource allocation strategy effectively encourage cooperation among clients, thereby improving FL performance. Future work will expand this approach to support a larger number of FL clients by incorporating explicit client selection at each round, along with spectrum and power allocation at each time slot.

\begin{appendices}
\section{Proof of Lemma 1} \label{proof: Lemma 1}
Local drift refers to the phenomenon where, during $E$ rounds of client training, the model parameters progressively deviate as mini-batch SGD is performed sequentially. This deviation introduces additional randomness into each computed gradient. Considering this drift, it can be demonstrated that cumulative stochastic gradient $\tilde{g}_{n,t}$ remains square-bounded. The detailed proof is presented as follows.
\begin{align}
		&~\mathbb{E}\left\| w_{n,t}^{(i)} - w_{t}\right\|^2 \notag \\
		&=\mathbb{E}\left\| w_{n,t}^{(i-1)} - w_{t} - \eta_\text{l}\left(\widetilde{\nabla}F_n(w_{n,t}^{(i)}) \pm \nabla F_n(w_{n,t}^{(i)}) \right)\right\|^2 \notag \\
		&\leq \mathbb{E}\left\| w_{n,t}^{(i-1)} - w_{t} - \eta_\text{l}\nabla F_n(w_{n,t}^{(i)}) \right\|^2 + \eta_\text{l}^2 \sigma_\text{l}^2 \label{proof:lemma 1-1} \\
		&=\mathbb{E}\left[ \left\| w_{n,t}^{(i-1)} - w_{t}\right\|^2 - 2 \eta_\text{l} \left \langle w_{n,t}^{(i-1)} - w_{t},F_n(w_{n,t}^{(i)})  \right\rangle \right. \notag\\
        &+ \eta_\text{l}^2 \left\| \nabla F_n(w_{n,t}^{(i)}) \right\|^2\bigg] + \eta_\text{l}^2 \sigma_\text{l}^2 \notag \\
		&= \mathbb{E}\left[ \left\| w_{n,t}^{(i-1)} - w_{t}\right\|^2 - 2 \eta_\text{l} \left \langle \frac{1}{\sqrt{2E-1}}\left(w_{n,t}^{(i-1)} - w_{t}\right), \right. \right. \notag \\
        & \quad \left.\left. \sqrt{2E-1}F_n(w_{n,t}^{(i)}) \right\rangle + \eta_\text{l}^2 \left\| \nabla F_n(w_{n,t}^{(i)}) \right\|^2 \right] + \eta_\text{l}^2 \sigma_\text{l}^2 \notag \\
		&\leq (1+\frac{1}{2E-1}) \mathbb{E} \left\| w_{n,t}^{(i-1)} - w_{t}\right\|^2 + 2 \eta_\text{l}^2 E ~ \mathbb{E} \left\| F_n(w_{n,t}^{(i)}) \right\|^2\notag \\
        & + \eta_\text{l}^2 \sigma_\text{l}^2 \label{proof:lemma 1-2} \\
		&\leq (1+\frac{1}{2E-1}) \mathbb{E} \left\| w_{n,t}^{(i-1)} - w_{t}\right\|^2 + 4\eta_\text{l}^2 E ~\mathbb{E} \left\| F_n(w_{n,t}^{(i)}) \right. \notag \\
        &- F_n(w_{t})\Big\|^2 + 4\eta_\text{l}^2 E \left\| F_n(w_{t})\right\|^2 + \eta_\text{l}^2 \sigma_\text{l}^2 \notag \\
		&\leq (1+\frac{1}{2E-1} + 4E\eta_\text{l}^2L^2)~ \mathbb{E} \left\| w_{n,t}^{(i-1)} - w_{t}\right\|^2 \notag \\
        &+ 4\eta_\text{l}^2 E \left\| F_n(w_{t})\right\|^2 + \eta_\text{l}^2 \sigma_\text{l}^2, \label{proof:lemma 1-3}
\end{align}
where \eqref{proof:lemma 1-1} follows from Assumption \ref{assumption:Unbiasedness and Bounded Local Variance}, \eqref{proof:lemma 1-2} is derived via arithmetic-geometric mean inequality, and \eqref{proof:lemma 1-3} results from Assumption \ref{assumption:L-Smoothness}.

When $E > 1$, if the client learning rate satisfies $\eta_\text{l} \leq \frac{1}{\sqrt{8}EL}$, then the inequality $1 + \frac{1}{E-1} \geq 1 + \frac{1}{2E-1} + 4E\eta_\text{l}^2L^2$ holds. Therefore,
\begin{align}
    \mathbb{E}\left\| w_{n,t}^{(i)} - w_{t}\right\|^2 &\leq \left(1+\frac{1}{E-1}\right) \mathbb{E} \left\| w_{n,t}^{(i-1)} - w_{t}\right\|^2 \notag \\
    &+ 4\eta_\text{l}^2 E \left\| F_n(w_{t})\right\|^2 + \eta_\text{l}^2 \sigma_\text{l}^2. \label{proof:lemma 1-4}
\end{align}	

Substituting $w_{n,t}^{(0)}=w_t$ into \eqref{proof:lemma 1-4} and solving recursively, we obtain that
\begin{align}
	&\!\!\!\!\!~\mathbb{E}\left\| w_{n,t}^{(i)} - w_{t}\right\|^2 \notag \\
    &\!\!\!\!\!\leq \sum_{p=1}^{i-1} \left(1+\frac{1}{E-1}\right)^p \eta_\text{l}^2 \left( 4E\left\| F_n(w_{t})\right\|^2 + \sigma_\text{l}^2 \right) \notag\\
	&\!\!\!\!\!=(E-1) \!\!\left[ \!\left(1+\frac{1}{E-1}\right)^i\!\! -\! 1\right] \!\eta_\text{l}^2\! \left( 4E\left\| F_n(w_{t})\right\|^2 \!\!+\! \sigma_\text{l}^2 \right)\!.\label{proof:lemma 1-5}
\end{align}

During local training process, the number of mini-batch SGD iterations is fixed at $E$. Substituting $i=E$ into \eqref{proof:lemma 1-5} and noting that the inequality $e < (1+\frac{1}{x-1})^x \leq 4$ holds for $x \geq 2$, we have
\begin{align}
	&\mathbb{E}\left\| w_{n,t}^{(E)} - w_{t}\right\|^2 \notag \\
    &\leq (E-1) \left[ \left(1+\frac{1}{E-1}\right)^E - 1\right] \eta_\text{l}^2 \left( 4E\left\| F_n(w_{t})\right\|^2 + \sigma_\text{l}^2 \right) \notag \\
	&\leq 3 \eta_\text{l}^2 E \left( 4E\left\| F_n(w_{t})\right\|^2 + \sigma_\text{l}^2 \right).
\end{align}

Finally, the proof of Lemma \ref{lemma:Local Drift} is thus completed, given by
\begin{align}
    \mathbb{E}\left\| w_{n,t}^{(i)} - w_{t}\right\|^2 &\leq \mathbb{E}\left\| w_{n,t}^{(E)} - w_{t}\right\|^2 = \mathbb{E}\left\| \tilde{g}_{n,t} \right\|^2 \notag\\
    &\leq 12 \eta_\text{l}^2 E^2\left\| F_n(w_{t})\right\|^2 + 3\eta_\text{l}^2 E \sigma_\text{l}^2.
\end{align}	
\vspace{-0.2cm}


\section{Proof of Lemma 2} \label{proof: Lemma 2}

Expanding the left-hand side of Assumption \ref{assumption:Bounded Global Variance}, we get
\vspace{-0.1cm}
\begin{equation*}
\begin{aligned}
    &~\frac{1}{N}\sum_{n=1}^N\left\|\nabla F_n(w)-\nabla F(w)\right\|^2 \\
	&=\!\frac{1}{N}\!\sum_{n=1}^N\!\left\|\nabla F_n(w)\right\|^2 \!\!- \! \frac{2}{N}\!\sum_{n=1}^N\!\!\left \langle \nabla F_n(w),\!\nabla F(w) \right\rangle \!+\! \left\|\nabla F(w)\right\|^2\!\!. 
\end{aligned}
\end{equation*}
\vspace{-0.2cm}

Substituting that $\nabla F(w) = \frac{1}{N}\sum_{n=1}^{N}\nabla F_n(w)$, we obtain
\begin{equation}
	\frac{1}{N}\sum_{n=1}^N\left\|\nabla F_n(w)\right\|^2 - \left\|\nabla F(w)\right\|^2 \leq \sigma_\text{g}^2.
\end{equation}
\vspace{-0.2cm}

Reorganizing, we conclude the proof of Lemma \ref{lemma:Bounded Average Squared Norm}, given by
\vspace{-0.2cm}
\begin{equation}
	\frac{1}{N}\sum_{n=1}^N\left\|\nabla F_n(w)\right\|^2 \leq \left\|\nabla F(w)\right\|^2+\sigma_\text{g}^2.
\end{equation}
\vspace{-0.5cm}

\section{Proof of Theorem 1} \label{proof: Theorem 1}

Considering Assumption \ref{assumption:L-Smoothness}, the one-step convergence upper bound of the FedAvg algorithm is given by
\begin{align}
	&\!\!\!\!\!\!\!\!\!~\mathbb{E} \left[F(w_{t+1})\right] - F(w_{t}) \notag\\
    &\!\!\!\!\!\!\!\!\!\leq \mathbb{E} \left \langle \nabla F(w_{t}), w_{t} - w_{t+1} \right\rangle + \frac{L}{2} \mathbb{E} \left\| w_{t} - w_{t+1}\right\|^2 \notag\\
	&\!\!\!\!\!\!\!\!\!= \!-\eta_\text{g} \mathbb{E} \left\langle\nabla F(w_{t}), \tilde{g}_{t} \right\rangle + \frac{L}{2} \eta_\text{g}^2 \mathbb{E} \left\|\tilde{g}_{t}\right\|^2 \notag \\
	&\!\!\!\!\!\!\!\!\!= \!-\eta_\text{g}\!\left\|\nabla F(w_t)\right\|^2 \!\!+ \!\eta_\text{g} \mathbb{E} \!\left\langle \nabla F(w_{t}), \!\nabla F(w_t)\! -\! \tilde{g}_{t} \right\rangle \!\!+ \!\frac{L}{2} \eta_\text{g}^2 \mathbb{E}\! \left\|\tilde{g}_{t}\right\|^2 \notag \\
	&\!\!\!\!\!\!\!\!\!\leq \!-\frac{\eta_\text{g}}{2}\!\left\|\nabla F(w_t)\right\|^2\!\! +\! \frac{\eta_\text{g}}{2}\underbrace{\mathbb{E}\left\| \nabla F(w_t) - \tilde{g}_{t}\right\|^2}_{\text{(A)}}\!+ \frac{L}{2} \eta_\text{g}^2 \underbrace{\mathbb{E} \left\|\tilde{g}_{t}\right\|^2}_{\text{(B)}}\!,\label{proof:theorem 1-2}
\end{align}
where \eqref{proof:theorem 1-2} is derived via arithmetic-geometric mean inequality. 

Firstly, by decomposing term (A), we obtain
\vspace{-0.15cm}
\begin{align}
	\allowdisplaybreaks
	&\!\!\!\!\mathbb{E}\left\| \nabla F(w_t) - \tilde{g}_{t}\right\|^2 \notag \\
    &\!\!\!\!= \mathbb{E}\left\| \nabla F(w_t) - \bar{g}_{t} + \bar{g}_{t} - \tilde{\bar{g}}_{t} + \tilde{\bar{g}}_{t} -\tilde{g}_{t}\right\|^2 \notag \\
	&\!\!\!\! \leq 2 \underbrace{\mathbb{E}\left\| \nabla F(w_t) - \bar{g}_{t} \right\|^2}_{(\text{A}_1)} + 2 \underbrace{\mathbb{E}\left\| \tilde{\bar{g}}_{t} -\tilde{g}_{t} \right\|^2}_{(\text{A}_2)} + \underbrace{\mathbb{E}\left\| \bar{g}_{t} - \tilde{\bar{g}}_{t}\right\|^2}_{(\text{A}_3)}. \notag
\end{align}
\vspace{-0.3cm}

Secondly, turning to term (B), we derive that
\vspace{-0.15cm}
\begin{align}
	\allowdisplaybreaks
	\mathbb{E} \left\|\tilde{g}_{t}\right\|^2 &= \mathbb{E}\left\| \tilde{g}_{t} -\tilde{\bar{g}}_{t} + \tilde{\bar{g}}_{t} \right\|^2 \notag \\
	& \leq 2 \underbrace{\mathbb{E}\left\| \tilde{\bar{g}}_{t} -\tilde{g}_{t} \right\|^2}_{(\text{A}_2)} + 2 \underbrace{\mathbb{E}\left\| \tilde{\bar{g}}_{t}  \right\|^2}_{(\text{A}_4)}.\notag
\end{align}
\vspace{-0.3cm}

After decomposing terms (A) and (B), we derive four terms. Notably, term $(\text{A}_2)$ depends on the specific selection of $\mathcal{N}_t$, which can inspire the design of the reward function in RL algorithms. 

Thirdly, by expanding term $(\text{A}_1)$, we obtain
\begin{align}
	\allowdisplaybreaks
		&~\mathbb{E}\left\| \nabla F(w_t) - \bar{g}_{t} \right\|^2 \notag \\
		&= \mathbb{E}\left\| \frac{1}{N}\sum_{n=1}^N \nabla F_n(w_t) - \frac{1}{N}\sum_{n=1}^N g_{n,t} \right\|^2 \notag \\
		&\leq \frac{1}{N} \sum_{n=1}^N \mathbb{E}\left\| \frac{1}{E}\sum_{i=1}^{E}\nabla F_n(w_t) - \eta_\text{l} \sum_{i=1}^{E}{\nabla F_n\left(w_{n,t}^{\left(i-1\right)}\right)} \right\|^2 \notag \\
		&= \frac{1}{NE^2} \sum_{n=1}^N \mathbb{E}\left\| (1-\eta_\text{l} E)\sum_{i=1}^{E}\nabla F_n(w_t) \right.\notag\\
        & \left.- \eta_\text{l} E \left(\sum_{i=1}^{E}\nabla F_n\left(w_{t}\right) -\sum_{i=1}^{E}{\nabla F_n\left(w_{n,t}^{\left(i-1\right)}\right)} \right)\right\|^2 \notag \\
		&\leq \frac{2}{N}(1-\eta_\text{l} E)^2 \sum_{n=1}^N \mathbb{E}\left\| \nabla F_n(w_t) \right\|^2 \notag \\
        & + \frac{2}{N} \eta_\text{l}^2 \sum_{n=1}^N \mathbb{E} \left\| \sum_{i=1}^{E} \left(\nabla F_n\left(w_{t}\right) -\nabla F_n\left(w_{n,t}^{\left(i-1\right)}\right)\right) \right\|^2 \label{proof:theorem A1-1}\\
		&\leq \frac{2}{N}(1-\eta_\text{l} E)^2 \sum_{n=1}^N \mathbb{E}\left\| \nabla F_n(w_t) \right\|^2 \notag \\
        &+ \frac{2}{N} \eta_\text{l}^2 \sum_{n=1}^N E \sum_{i=1}^{E}\mathbb{E} \left\|  \nabla F_n\left(w_{t}\right) -\nabla F_n\left(w_{n,t}^{\left(i-1\right)}\right) \right\|^2 \label{proof:theorem A1-2}\\
		&\leq \frac{2}{N}(1-\eta_\text{l} E)^2 \sum_{n=1}^N \mathbb{E}\left\| \nabla F_n(w_t) \right\|^2 \notag \\
        &+ \frac{2}{N} \eta_\text{l}^2 E L^2\sum_{n=1}^N \sum_{i=1}^{E}\mathbb{E} \left\|  w_{t} - w_{n,t}^{\left(i-1\right)} \right\|^2 \label{proof:theorem A1-3} \\
		&\leq \frac{2}{N}(1-\eta_\text{l} E)^2 \sum_{n=1}^N \mathbb{E}\left\| \nabla F_n(w_t) \right\|^2 \notag \\
        &+ \frac{2}{N} \eta_\text{l}^2 E  L^2 \sum_{n=1}^N \sum_{i=1}^{E}3E\eta_\text{l}^2 \left( 4E\left\| F_n(w_{t})\right\|^2 + \sigma_\text{l}^2 \right) \label{proof:theorem A1-4}\\
		&= \frac{2}{N}(1-\eta_\text{l} E)^2 \sum_{n=1}^N \mathbb{E}\left\| \nabla F_n(w_t) \right\|^2 \notag \\
        &+ \frac{24}{N} \eta_\text{l}^4 E^4 L^2 \sum_{n=1}^N  4E\left\| F_n(w_{t})\right\|^2 + 6\eta_\text{l}^4 E^3 L^2\sigma_\text{l}^2, \notag\\
        &\leq \left[ 2 (1-\eta_\text{l} E)^2 + 24 \eta_\text{l}^4 E^4 L^2  \right] \left\| \nabla F(w_t) \right\|^2 \notag \\
        &+ 6\eta_\text{l}^4 E^3 L^2 \sigma_\text{l}^2 +\left[ 2 (1-\eta_\text{l} E)^2 + 24 \eta_\text{l}^4 E^4 L^2 \right] \sigma_\text{g}^2. \label{proof:A1-final}
\end{align}
where \eqref{proof:theorem A1-1} follows from the arithmetic-geometric mean inequality, \eqref{proof:theorem A1-2} is derived using the Cauchy inequality, \eqref{proof:theorem A1-3} relies on the Assumption \ref{assumption:L-Smoothness} and \eqref{proof:theorem A1-4} is deduced from from Lemma \ref{lemma:Local Drift}, \eqref{proof:A1-final} is derived using Lemma \ref{lemma:Bounded Average Squared Norm}.

Fourthly, considering term $(\text{A}_3)$, we substitute the definitions of $\bar{g}_{t}$ and $\tilde{\bar{g}}_{t}$, yielding
\begin{align}
		&~\mathbb{E}\left\| \bar{g}_{t} - \tilde{\bar{g}}_{t}\right\|^2 \notag \\
        &=\mathbb{E}\left\| \frac{1}{N}\sum_{n=1}^{N}{g}_{n,t} - \frac{1}{N}\sum_{n=1}^{N}{\tilde{g}}_{n,t} \right\|^2 \notag \\
		&= \frac{1}{N^2} \sum_{n=1}^{N} \mathbb{E}\left\| \eta_\text{l} \sum_{i=1}^{E}{\left( \nabla F_n\left(w_{n,t}^{\left(i-1\right)}\right) - \widetilde{\nabla}F_n\left(w_{n,t}^{\left(i-1\right)}\right) \right)} \right\|^2 \notag \\
		&=\frac{\eta_\text{l}^2}{N^2} \sum_{n=1}^{N} \sum_{i=1}^{E} \mathbb{E} \left\|\nabla F_n\left(w_{n,t}^{\left(i-1\right)}\right) - \widetilde{\nabla}F_n\left(w_{n,t}^{\left(i-1\right) } \right) \right\|^2\notag \\
		&\leq \frac{\eta_\text{l}^2}{N^2} \sum_{n=1}^{N} \sum_{i=1}^{E} \sigma_\text{l}^2 \label{proof:theorem A3-1} \\
		&= \frac{\eta_\text{l}^2 E^2 \sigma_\text{l}^2}{N},\notag 
\end{align}
where \eqref{proof:theorem A3-1} is derived from Assumption \ref{assumption:Unbiasedness and Bounded Local Variance}.

Finally, by expanding term $(\text{A}_4)$ and substituting the definition of $\tilde{\bar{g}}_{t}$, we derive that
\begin{align}
    \allowdisplaybreaks
	&\mathbb{E}\left\| \tilde{\bar{g}}_{t} \right\|^2 =  \mathbb{E}\left\| \frac{1}{N}\sum_{n=1}^{N}\tilde{g}_{n,t}  \right\|^2 \notag \\
	&\leq \frac{1}{N}\sum_{n=1}^{N} \mathbb{E}\left\| \tilde{g}_{n,t}  \right\|^2  \notag\\
	&\leq \frac{1}{N}\sum_{n=1}^{N} 12 \eta_\text{l}^2 E^2\left\| F_n(w_{t})\right\|^2 + 3\eta_\text{l}^2 E \sigma_\text{l}^2 \label{proof:theorem A4-2}\\
	&\leq 12 \eta_\text{l}^2 E^2 \left\| \nabla F(w_t) \right\|^2 + 12 \eta_\text{l}^2 E^2 \sigma_\text{g}^2 + 3\eta_\text{l}^2 E \sigma_\text{l}^2, \label{proof:theorem A4-3}
\end{align}
where \eqref{proof:theorem A4-2} is inferred from Lemma \ref{lemma:Local Drift}, and \eqref{proof:theorem A4-3} is deduced from Lemma \ref{lemma:Bounded Average Squared Norm}.

Integrating the preceding derivations, we finalize the proof of Theorem \ref{theorem:One-Step Convergence}, yielding the one-step convergence bound as
\begin{equation}
		\mathbb{E}[F(w_{t+1})] - F(w_{t}) \leq C_1 \|\nabla F(w_{t})\|^2 + C_2 \mathbb{E} \|\tilde{\bar{g}}_{t} -\tilde{g}_{t}\|^2 + C_3,
\end{equation}
where
\begin{equation*}
	\begin{aligned}
		C_1 &= 2 \eta_\text{g} (1 - \eta_\text{l} E)^2  - \frac{1}{2} \eta_\text{g} + 12 \eta_\text{g}^2 \eta_\text{l}^2 E^2 L+ 24 \eta_\text{g} \eta_\text{l}^4 E^4 L^2, \\
		C_2 &= \eta_\text{g} (1 + \eta_\text{g} L),\\
		C_3 &= \left[2 \eta_\text{g} (1 - \eta_\text{l} E)^2  + 12 \eta_\text{g}^2 \eta_\text{l}^2 E^2 L + 24 \eta_\text{g} \eta_\text{l}^4 E^4 L^2\right] \sigma_\text{g}^2 \\
        &+ (\frac{1}{2N} \eta_\text{g} \eta_\text{l}^2 E^3 + 3 \eta_\text{g}^2 \eta_\text{l}^2 E L + 6 \eta_\text{g} \eta_\text{l}^4 E^3 L^2)\sigma_\text{l}^2.
	\end{aligned}
\end{equation*}

\end{appendices}

{\small
\bibliographystyle{IEEEtran}  
\bibliography{journal_dc.bib}  
}

\end{document}